\documentclass[review]{elsarticle}

\usepackage{lineno,hyperref}
\modulolinenumbers[5]

\usepackage{graphicx}
\usepackage{fullpage}
\usepackage{amsmath}
\usepackage[algo2e]{algorithm2e} 
\usepackage{algorithm}
\usepackage{algorithmicx}

\usepackage[utf8]{inputenc}
\usepackage[english]{babel}
\newtheorem{lemma}{Lemma}

\newtheorem{theorem}{Theorem}
\newtheorem{definition}{Definition}

\journal{Journal of Parallel and Distributed Computing}

\begin{document}

\begin{frontmatter}

\title{Uniform Circle Formation By Oblivious Swarm Robots}





\author {Moumita Mondal$^1$, Sruti Gan Chaudhuri$^1$, Ayan Dutta$^2$, Krishnendu Mukhopadhyaya$^3$, Punyasha Chatterjee$^1$}

\address{$^1$ Jadavpur University, Kolkata, India, $^2$ University of North Florida, Jacksonville, USA, $^3$ Indian Statistical Institute, Kolkata, India}

\begin{abstract}
In this paper, we study the circle formation problem by multiple autonomous and homogeneous disc-shaped robots (also known as fat robots). The goal of the robots is to place themselves on the periphery of a circle. Circle formation has many real-world applications, such as boundary surveillance. This paper addresses one variant of such problem -- uniform circle formation, where the robots have to be equidistant apart. The robots operate by executing cycles of the states `wait-look-compute-move'. They are oblivious, indistinguishable, anonymous, and do not communicate via message passing. First, we solve the uniform circle formation problem while assuming the robots to be transparent. Next, we address an even weaker model, where the robots are non-transparent and have limited visibility. We propose novel distributed algorithms to solve these variants. Our presented algorithms in this paper are proved to be correct and guarantee to prevent collision and deadlock among the swarm of robots.
\end{abstract}


\end{frontmatter}


\section{Introduction}
One of the current trends of research in the field of robotics is to replace costly robots having multiple sensors by a group of tiny autonomous robots who work in coordination between themselves. These robots are basically programmable particles and popularly known as {\it swarm robots}. The goal of this mobile robot system is often to perform patrolling, sensing, and exploring in a harsh environment such as disaster area, deep sea and space without any human intervention \cite{Ref7}. Theoretical representation of such mobile robot systems in the two-dimensional Euclidean space attracts much attention where one of the fundamental tasks for executing this kind of job is to form the geometric patterns on the plane by the robots' positions. The investigation on distributed or decentralized control of these robots with very limited capabilities is an emerging area of research at present. In order to execute some jobs in collaboration by the robots, it may require some geometric shapes or patterns to be formed by the positions of the robots on a 2D plane. The significance of positioning the robots based on some given patterns may be useful for various tasks, such as operations in hazardous environments, space missions, military operations, tumor excision, etc \cite{Ref7}. This paper addresses one such geometric problem, uniform circle formation, where the robots place themselves equidistant apart on the boundary of a circle. Each robot is capable of sensing its immediate surrounding (i.e nearby robots), performing computations on the sensed data, and moving towards the computed destination. A significant application of the uniform circle formation algorithm, during this Covid-19 pandemic, maybe bordering a region with autonomous robots, having UV ray emitting capabilities. The boundary can be disinfected by the emission of UV light \cite{Ref21} \cite{Ref22} \cite{Ref23}. The robots are free to move on a 2D plane, and are anonymous, oblivious and can only interact by observing others position. Based on this model, we study the problem of uniform circle formation by fat {\it (i.e. unit-disc) } swarm robots with different visibility ranges. The first part, ({\it i.e., section 3}) presents a distributed algorithm with swarm robots having unlimited visibility. The second part ({\it i.e., section 4}) of the paper proposes another distributed algorithm with limited visibility swarm robots, i.e., the robots can observe other robots around itself within a fixed distance. The visibility range is the same for all the robots. Also, the robots can be inside or outside the given circle. Finally the last part ({\it i.e., section 5})  of the paper discusses the second distributed algorithm for uniform circle formation by swarm robots with non-uniform visibility ranges. Here the visibility range may be different for the different robots.
\subsection{Framework}
The world of swarm robots consists of multiple mobile robots moving on a 2D plane\cite{Ref5}. The robots consist of (i) Motorial capabilities i.e., they can independently move in a Euclidean space; and (ii) Sensorial capabilities i.e., they sense the locations of the other robots. However, the robots have no explicit way of communication. The robots coordinate among themselves by observing the positions of the other robots on the plane. A robot is always able to see another robot within its visibility range (limited or unlimited). Additionally, the robots are {\it homogeneous} (all executes the same algorithm) and {\it anonymous} (no unique identifiers). The {\it autonomy} of the robots system allows the robots to work without centralized control. The robots are assumed to be correct or non-faulty. The robots are considered to be unit-discs or fat robots such that the radius of a disc robot is unit distance. They act as physical obstructions for other robots. We represent a robot by its center. The robots do not stop before reaching its destination, this movement is known as {\it rigid} movement. The robots are oblivious, i.e. they do not carry forward any information from their previous computational cycles. In general, the robots do not have any common coordinate system and orientation.
The robots execute a cycle of four phases:
\begin{itemize}
\item {\bf Wait} - The robots may be inactive or idle in the wait state.
\item {\bf Sense} - In this state, a robot takes a snapshot of its surroundings, within its range of vision.
\item {\bf Compute} - In the compute state, it executes an algorithm for computing the destination to move to. The algorithm is the same for all robots.
\item {\bf Move} - The robot moves to the computed destination.
\end{itemize}
The robots may or may not be synchronized.
\begin{itemize}
\item In {\bf fully-synchronous (FSYNC)} model, all the robots execute their cycles together. In such a system, all robots get the same view. As a result, they compute on the same data.
\item In {\bf semi-synchronous (SSYNC)} model, a set of robots execute their cycles together. Under this scheduling, there is a global clock but, at each cycle, a robot may or may not be active. This scheduling assures that when a robot is moving no other robot is sensing.
\item A more practical model is {\bf asynchronous (ASYNC)} model, where the actions of the robots are independent. By the time a robot completes its computation, several of the robots may have moved from the positions based on which the computation is made. Here, a robot in motion is visible. 
\end{itemize}
This model is known as CORDA (Co-OpeRative Distributed Asynchronous) model \cite{Ref24}.
We propose the robot's move strategy such that, after a finite time they are placed equidistantly apart on a boundary of a circle forming a uniform circle or regular polygon.

\section{Earlier Works and Our Contribution}
The problem of Circle Formation by mobile robots has been investigated by many researchers \cite{Ref3}\cite{Ref4}\cite{Ref17}\cite{Ref18}\cite{Ref19} \cite{Ref20}.
A large body of research work exists in the context of multiple autonomous swarm robots exhibiting cooperative behavior. Such research aims to study the issues as group architecture, resource conflict, origin of cooperation, learning and geometric problems \cite{Ref9}. Traditional or conventional approach to swarm robots involves artificial intelligence in which most of the results are based on experimental study or simulations. Recently a new emerging field of robot swarm looks at the robots as distributed mobile entities and studies several coordination problems and proceed to solve them deterministically providing proof of correctness of the algorithms. 
The computational model popular in the literature under this field for mobile robots is called a {\it weak model} \cite{Ref5}. Here, the robots execute a cycle consisting of four phases, wait-look-compute-move. The robots do not communicate through any wired or wireless medium. The robots may execute the cycle synchronously or semi-synchronously or asynchronously.

Sugihara and Suzuki \cite{Ref8} proposed a simple heuristic algorithm for the formation of an approximation of a circle under limited visibility. Flocchini et.al \cite{Ref10}, solved the uniform circle formation problem for anonymous, autonomous, oblivious, disoriented point robots. It has been proved that the Uniform Circle Formation problem is solvable for any initial configuration of n$\ne$4 robots without any additional assumption. Recently, Mamino and Viglietta \cite{Ref11} solved the uniform circle formation for four point robots, thus completing the problem of uniform circle formation for any initial configuration for point robots without any extra assumption. All of these algorithms assume the robots to be a point which neither creates any visual obstruction nor acts as an obstacle in the path of other robots. Obviously such robots are not practical. However small they might be, they must have certain dimensions. 

Czyzowicz et al.,\cite{Ref2} extend the traditional weak model \cite{Ref5} of robots by replacing the point robots with unit disc robots. They named these robots as fat robots. Gan Chaudhuri and Mukhopadhyaya \cite{Ref6} proposed an algorithm for gathering multiple fat robots. Many of the previous circle formation algorithms required the system to be synchronous which is also an ideal situation. Most of the earlier works considered that the robots have unlimited visibility range, i.e., a robot can see infinite radius around itself.

Ando et al. \cite{Ref1} proposed a point convergence algorithm for oblivious robots in limited visibility. Later Flocchini et al. presented a gathering algorithm for asynchronous, oblivious robots in limited visibility having a common coordinate system. 
Souissi et al. \cite{Ref11} studied the solvability of gathering in limited visibility for the semi-synchronous model of robots using an unreliable compass. They assume that the compass is unstable for some arbitrary long periods and stabilize eventually. 
Dutta et. al \cite{Ref15}\cite{Ref12} have proposed  circle formation algorithms for fat robots. Recently,  R. Yang, \cite{Ref14} et al. has reported a simulated based result on  uniform circle formation of fat robots under limited visibility range where the sensing range of all robots are equal. 

Any result on {\it uniform} circle formation for fat robots (considering the distributed model) under different visibility ranges are not yet reported. In this paper, we propose distributed algorithms to form a {\it convex regular polygon} in other words a {\it uniform circle} by fat robots with different visibility capabilities. The uniform circle formation algorithms proposed in the later sections of the paper are influenced by the non-uniform circle formation algorithm presented in \cite{Ref12}. In \cite{Ref12}, the author has also assumed that initially all robots lie inside the target circle. We have removed this assumption; i.e., we  have no restriction on the initial configuration of the robots.

\subsection{Our Contribution}
To the best of our knowledge, this is the first work that address uniform circle formation by oblivious fat robots with different visibility ranges. Our primary contributions in this paper are:
\begin{itemize}
\item In the first part of the paper we propose a distributed algorithm to form a uniform circle by fat robots with unlimited visibility. One of the main concerns in this algorithm is to avoid collisions among the robots. We show that if the robots are semi-synchronous, execute rigid motion and agree on only one axis (e.g., $Y$-axis), then they can form a uniform circle without encountering any collision.
\item In the second part of the paper, another distributed algorithm to form a {\it uniform circle} by fat robots with limited visibility is proposed. The algorithm works with asynchronous fat robots, that agree upon a common origin and axes.
\item In the third section, we modify the previous algorithm to form a {\it uniform circle} by fat robots with non-uniform limited visibility.  We show that if a group of opaque, asynchronous, fat robots have different radius visibility circles but the common origin, common axes and common sense of direction then they can form a uniform circle without encountering any collision and deadlock. 
\end{itemize}

\section{Uniform Circle Formation By Swarm Robots Under Unlimited Visibility}
A set of $n$ stationary points on the 2D plane, representing the centers of the fat robots is given. Every robot acts as a physical obstacle for the other robots. The robots move in such a way that after a finite number of cycle execution they are placed equidistantly apart on a circumference of a circle.  
The number of robots, $n>1$ and a length $a>3$ is given as the inputs of the algorithm. The distance between two adjacent robots on the uniform circle will be atleast $a$. Note that $a$ can also be interpreted as the minimum required length of the edges of a convex polygon constructed by the robots on the uniform circle. The robots do not explicitly agree on any unit distance.  Since, the robots are unit discs, implicitly the radius of the robots can be considered as a unit. Hence, the robots can agree on the length $a$ \footnote{$a$ can also be computed inside a different subroutine.}. The objective of the algorithm is to form a circle where the robots are placed equidistant apart on the circumference of this circle.

\subsection{Underlying Model}
Let R = {$r_{1}, r_{2},..,r_{n}$} be a set of unit disc-shaped autonomous robots referred as {\it fat robots}. A robot is represented by its center i.e., by $r_{i}$ we mean a robot whose center is $r_{i}$. The set of robots R deployed on the 2D plane is described as follows:
\begin{itemize}
    \item The robots are autonomous.
    \item Robots are anonymous and homogeneous.
    \item The robots are oblivious in the sense that they can not recollect any data from the past cycle.
    \item The robots have rigid movement.
    \item Robots can not communicate explicitly. Each robot is allowed to have a camera that can take pictures over 360 degrees. The robots communicate only by means of observing other robots with the camera.
     \item The robots are unit-disc (i.e. fat robots)
    \item The robots are transparent, but they act as obstacle for other robots.
     \item The robots have unlimited visibility.
     \item Each robot executes a cycle of wait-look-compute-move semi-synchronously.
     \item The robots do not have any common coordinate system and orientation. We assume that they only agree on the $Y$-axis. However, the direction of the $X$-axis is not the same for all the robots.
\end{itemize}

\subsection{Overview of the Problem}
A set of robots, R is given. Our objective is to form a uniform circle of unit disc robots with unlimited visibility, by moving the robots to the circumference of the SEC.
Following are the steps to be executed by each robot in the compute phase: 
\begin{itemize}
\item The robots compute the radius ($rad_{req}$) of the circle to be formed using the {\it ComputeRadius} routine. ({\it section 3.2})
\item The robots compute $rad$, the radius of the current Smallest Enclosing Circle (SEC)\footnote{The circle with minimum radius such that all the robots are either inside the circle or on the circle.} of $n$ robots.
\item If $rad < rad_{req}$, then a routine for expanding the SEC, {\it SECExpansion} is called. ({\it section 3.3})
\item Else, a routine for forming a uniform circle, {\it FormUCircle} is called. ({\it section 3.4}) 	
\end{itemize}
\paragraph{\textbf{Notations:}} The following notations are used throughout the paper:
\begin{itemize}
\item SEC: Smallest Enclosing Circle
\item $n$: Number of robots in R.
\item R: Set of the unit-disc robots; R = {$r_{1}, r_{2},..,r_{n}$}
\item $\{T_0, T_1, \ldots, T_n \}$: Equidistant target positions on the circumference of CIR.
\item $a$: Input length, i.e. the minimum distance between two adjacent robots on the circle ($a>3$).
\item $\alpha$: is the central angle of the arc in degrees, ($\alpha = \frac{360}{n}$)
\item $rad$: Radius of the current SEC.
\item $rad_{req}$: Radius of the required SEC, to accommodate all the n unit-disc robots on the SEC.
\item $d_R = 2(rad_{req}-rad)$
\item $L$: A line parallel to the Y-axis and passing through the center,c of the SEC.
\item $o$: The north-most intersection point of L and the SEC.
\end{itemize}
\subsection{Description of The Algorithm ComputeRadius}
Let the minimum radius of the SEC required to accommodate all $n$ fat robots be $rad_{req}$. The required minimum distance between two adjacent robots on the circle is given as $a$.  When there is no gap between two adjacent robots on the circle, the distance between the centers of two adjacent robots on the circle will be $2$ units. We assume that, $a$ is at least $3$ units in length. The algorithm ComputeRadius() finds the minimum radius of the circle to be formed.

\begin{figure}[H]
\centering
{\includegraphics[scale = 0.5,clip,keepaspectratio]{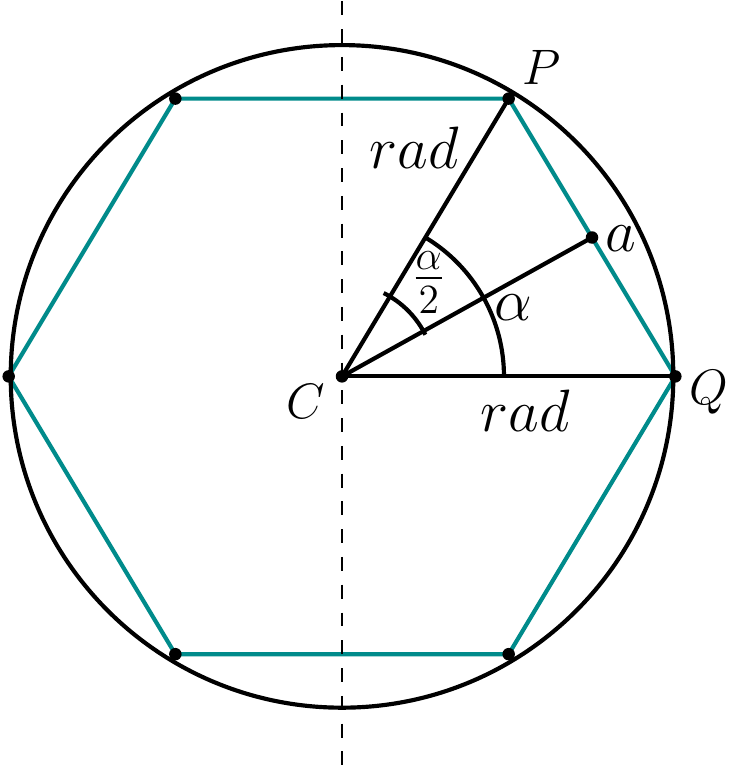}} 
\caption{Minimum radius of SEC required to accommodate all the $n$ robots.}
\label{i_5}
\end{figure}

Refer to Figure \ref{i_5}, 
$|PQ| = a$ and $PC=rad_{req}$. 
Hence,
\begin{equation}
\label{rad_{req}}
rad_{req}= \frac{a}{2sin(\frac{360}{2n})}
\end{equation}

\begin{algorithm}[H]
\KwIn{a, n}
\KwOut{$rad_{req}$}
return $rad_{req}= \frac{a}{2sin(\frac{360}{2n})}$\\
\caption{ComputeRadius(a,n)}
\normalsize
\end{algorithm}

\subsection{Description of The Algorithm SECExpansion}
Next the robots compute the radius of current SEC, $rad$. If this current SEC cannot accommodate all the $n$ fat robots (i.e. $rad < rad_{req}$), then the robots on SEC will move away from the center of SEC in order to expand it. Our algorithm does not allow all the robots on the SEC to move outside simultaneously. Instead one or two leader robots are selected who moves to expand the SEC. To assure collision-free movements the robots always move along {\it free paths} described as follows.

\begin{figure}[H]
   \centering
   \includegraphics[scale=0.52]{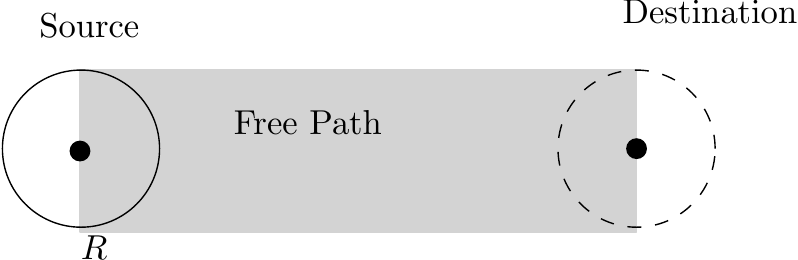}
   \caption{An example of a free path of robot R}
   \label{i_freepath}
\end{figure}

\begin{definition}
Free path is a path of a robot from source to destination point (Figure \ref{i_freepath}) such that, the rectangular area having the length as the source to destination distance and width as two units, does not contain any part of another robot. 

\end{definition}

\begin{figure}[H]
     \centering
   \includegraphics[scale=0.45]{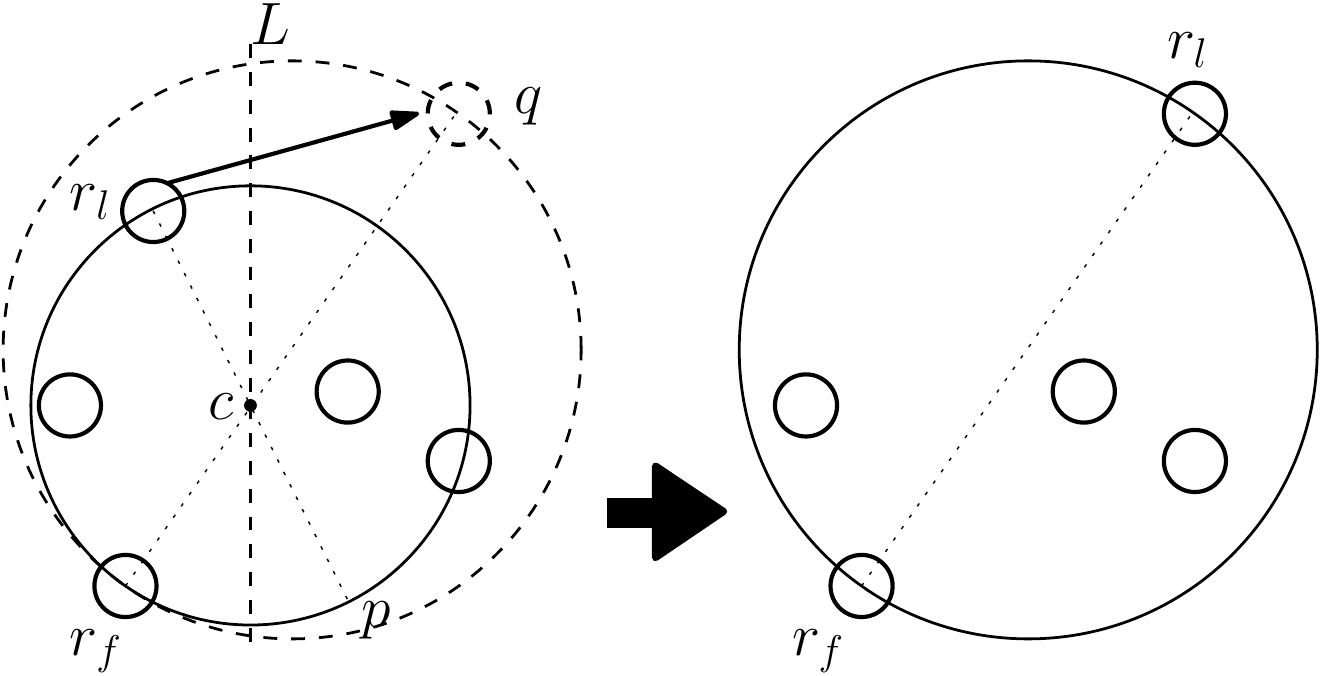} 
   \caption{The robots are in case 1 and there exists no robot at $p$.}
    \label{i_3}
\end{figure}

\begin{figure}[H]
   \centering
   \includegraphics[scale=0.42]{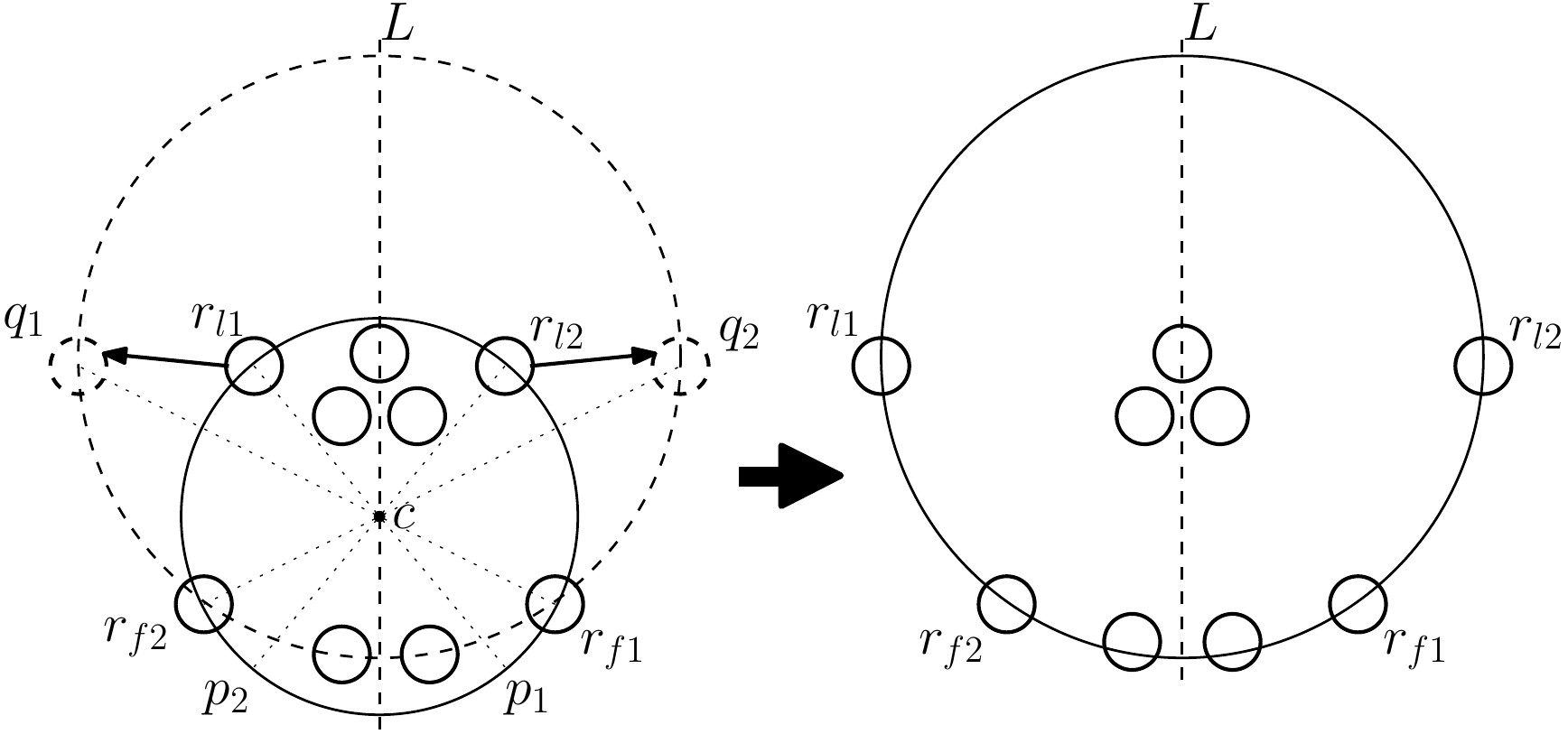}
   \caption{The robots are in case 2 and there exists no robot at $p_{i} (i=1,2)$ }
   \label{i_4}
\end{figure}

For any robot $R$, if $R$ lies on the circumference of the SEC, it moves following the below procedure, SECExpansion. 
\begin{itemize}
\item Under this procedure first a leader robot is elected. The robot which has a maximum Y value on the SEC  and has some unique feature is selected as the leader. Let $L$ be the line parallel to the Y-axis and passing through the center $c$ of the SEC. Three cases are possible.	
    \begin{itemize}
	\item {\bf Case 1.} The robot positions are not symmetric with respect to the line $L$. In this case, there exists a leader robot $r_l$ which has no mirror image robot with respect to $L$. 	
	\item {\bf Case 2.} The robot positions are symmetric with respect to $L$.	If the robots are in case 2, then there exist two leaders $r_{l1}$ and $r_{l2}$ on the SEC, which are the mirror images of each other. Note that these leaders have a maximum Y value. If there is any robot on the SEC, which has a maximum Y value and is on the intersection point of $L$ and the SEC, then it is not selected as the leader robot.
    \end{itemize}
\item If the robots are in case 1, then draw a line $r_lc$ ($c$: center of the SEC). Let $r_lc$ intersect the SEC at $p$.  
If the robots are in case 2, then draw lines  $r_{l1}c$ and $r_{l2}c$. Let $r_{l1}c$ intersect the SEC at $p1$ and $r_{l2}c$ intersect the SEC at $p2$.
\item If the robots are in case 1 and there exists a robot, $r_p$ at $p$ (Figure \ref{i_1}), then $r_l$ moves $d_R$ distance, radially away from $c$ where $d_R = 2(rad_{req}-rad)$.
If the robots are in case 2 and there exists a robot  $r_{pi}$ at $p_{i} (i=1,2)$ (Figure \ref{i_2}), then $r_{li} (i=1,2)$ moves $d_R$ distance, radially away  from $c$ where $d_R = 2(rad_{req}-rad)$. Due to semi-synchronous scheduling both $r_{l1}$ and $r_{l2}$ may or may not move simultaneously. However, if any of the robot moves, the SEC becomes as big as required. The center of the SEC moves (i) along the diameter $r_lc$ in case 1 and case 2 when one of the leaders moves or (ii) along $L$ in case 2, if both the leaders move.

\begin{figure}[H]
      \centering
   \includegraphics[scale=0.43]{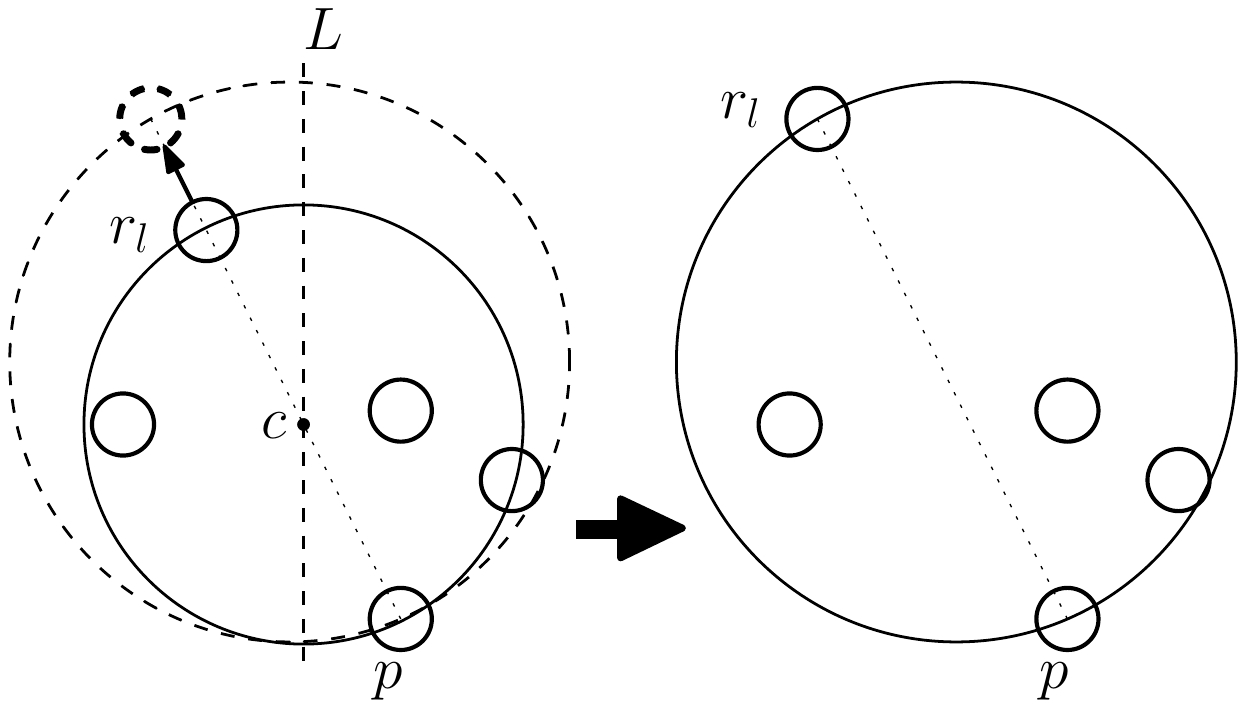} 
   \caption{The robots are in case 1 and there exists a robot at $p$.}
   \label{i_1}
   \end{figure}
 
 \begin{figure}[H]
   \label{2}
  \centering
   \includegraphics[scale=0.43]{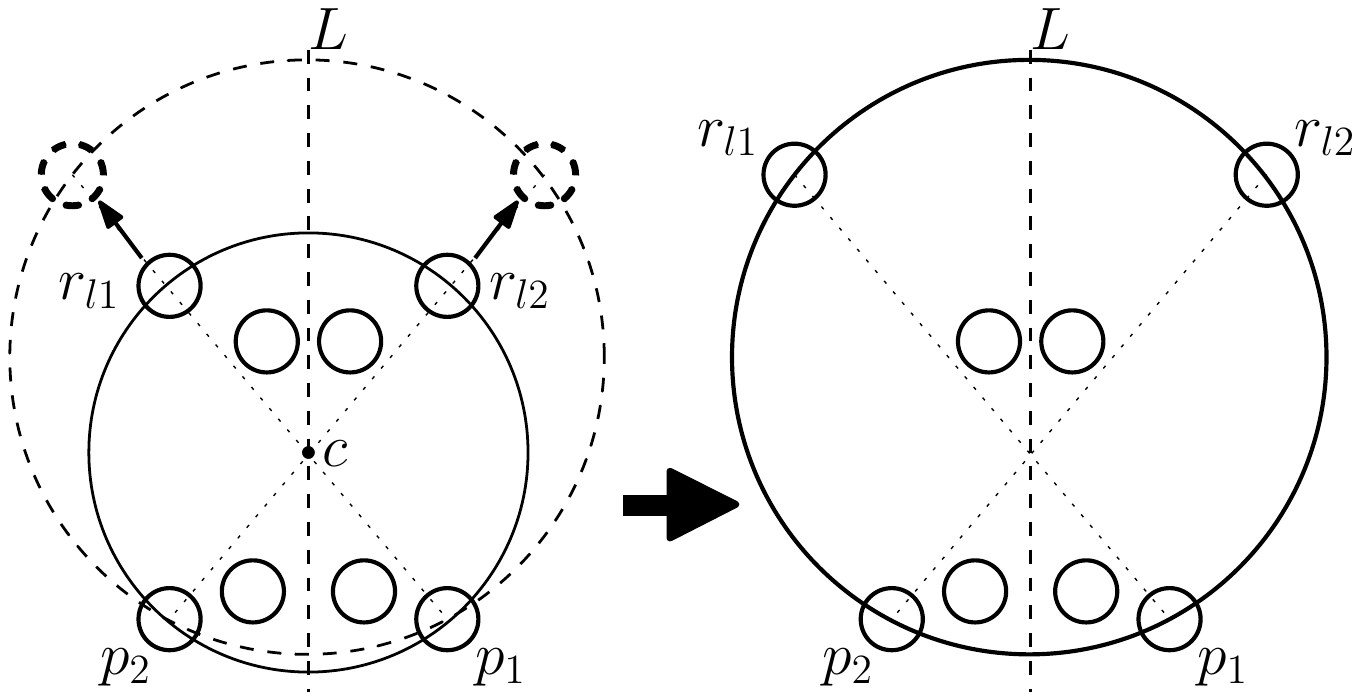}
   \caption{The robots are in case 2 and there exists a robot at $p_{i} (i=1,2)$ }
   \label{i_2}
   \end{figure}

\item If the robots are in case 1 and there exists no robot at $p$ (Figure \ref{i_3}), then let $r_f$ be a robot on the SEC which is farthest from $r_l$. Draw a line $r_fc$. Compute a point $q$ on the ray ${r_fc}$, such that $|r_fq|= 2rad$.
\begin{itemize}
\item If the path from $r_l$ to $q$ is a free path then $r_l$ moves to $q$.
\item Else, let $r'_l$ be the robot nearest to $q$, and has a free path to $q$. $r'_l$ moves to $q$. 
\end{itemize}
If the robots are in case 2 and there exists no robot at $p_{i} (i=1,2)$ (Figure \ref{i_4}), then let $r_{fi}(i=1,2)$ be the robots on the SEC which is farthest from $r_{li}(i=1,2)$. Draw the lines $r_{fi}c (i=1,2)$. Compute points $q_i(i=1,2)$ on the rays $r_{fi}c(i=1,2)$, such that $|r_fq_i|(i=1,2)= 2rad$.

\begin{itemize}
\item If the path from $r_{li}(i=1,2)$ to $q_i(i=1,2)$ is a free path then $r_{li}(i=1,2)$ move(s) to $q_i(i=1,2)$.
\item Else, let $r'_{li}(i=1,2)$ be the robot nearest to $q_i(i=1,2)$, and has a free path to $q_i(i=1,2)$. $r'_{li}(i=1,2)$ move(s) to $q_i(i=1,2)$. 
\end{itemize}
\end{itemize}

\begin{algorithm}[H]
\KwIn{$rad_{req}$.}
\KwOut{Expanded SEC with radius $rad_{req}$.}
Compute the SEC of $n$ robots;\\
$c=$ center of the SEC;
$rad=$radius of the SEC;
$s = r's$ location\\
\eIf{$s$ is inside the SEC}
{$d_R = 0$;}
{Compute the $Y$ value of all robots positions on the circumference of the SEC\\
\eIf{$Y_s$ (Y value of $s$) is maximum among all robots}
{ extend $sc$ to intersect the SEC at $p$\\
\eIf{there exists a robot $r_p$ at $p$}
{ $d_R = 2(rad_{req}-rad)$;
$R$ moves $d_R$ distance away from $c$ along $ps$; 
}
{ Let $r_f$ at $f$ be the robots farthest from $s$;\\
Extend $fc$ to $t$ such that $|ft|=rad_{req}$;
\eIf{$st$ is a free path}
{$ d_R = |st|$;
$R$ moves $d_R$ distance away from $c$ along $st$;
}
{Let $r_q$ be the robot at $q$  nearest to $t$ and $qt$ is a free path.\\
\eIf{$R=r_q$}
{$d_R = |qt|$;\\
$R$ moves $d_R$ distance away from $c$ along $qt$;
}
{$d_R = 0$;}
}
}
}
{$d_R = 0$;}
}
\caption{SECExpansion($rad_{req}$)}
\end{algorithm}


\begin{lemma}
When there exists a robot at $p$ in case 1 or there exists a robot at $p_{i} (i=1,2)$ in case 2, and when $r_l$ (case 1) or $r_{li} (i=1,2)$ (case 2) is moving outwards, then $r_l$ (case 1) or $r_{li} (i=1,2)$ (case 2) and $r_p$ (case 1) or $r_{pi} (i=1,2)$ (case 2) always remain on the current SEC.
\end{lemma}
\textbf{Proof:}
First consider case 1. Initially $r_l$ and $r_p$ are on the SEC and they are diagonally opposite to each other. Hence, $r_p$ is in maximum distance from $r_l$. No robot other than $r_l$  is moving. $r_l$ is moving following the straight line $r_pr_l$ and away from $r_p$. Thus the distance between $r_p$ and $r_l$ is increasing. $r_p$ continues to remain in maximum distance from $r_l$. According to the SEC property the maximum distant points of a point set lie on the SEC of that point set. Hence, $r_l$ and $r_p$ remains on the current SEC (or in the changing SEC). 
Now consider case 2. If any of $r_{li1}$ or $r_{li2}$ is moving, the case is similar to case 1. Otherwise, initially, $r_{pi} (i=1,2)$ lie at diagonally opposite of $r_{li} (i=1,2)$. hence, $r_{pi} (i=1,2)$ is in maximum distance from $r_{li}(i=1,2)$. No robot other than $r_{li}(i=1,2)$  is moving. $r_{li}(i=1,2)$ is moving following the straight line $r_{pi}r_{li}$ and away from $r_{pi}$. Thus the distance between $r_{pi}$ and $r_{li}$ (for $(i=1,2)$) is increasing. $r_{pi}(i=1,2)$ continues to remain in maximum distance from $r_{li}(i=1,2)$. According to the SEC property the maximum distant points of a point set lie on the SEC of that point set. Hence, $r_{li}$ and $r_{pi}$ remains on the current SEC (or in the changing SEC).    

\begin{lemma}
\label{gerc0}
When there exists a robot at $p$ in case 1 or there exists a robot at $p_{i} (i=1,2)$ in case 2, and when $r_l$ (case 1) or $r_{li} (i=1,2)$ (case 2) reaches its destination, then the radius of the new SEC is $\ge rad_{req}$.
\end{lemma}
\textbf{Proof:}
Consider case 1. $r_l$ and $r_p$ in a new position is the diameter of the new SEC. $r_l$ to a distance $2(rad_{req}-rad)$. Hence, the length of the diameter of this SEC is  $2rad$, i.e., the radius is $rad_{req}$.
Now consider case 2. Suppose, any one leader is moving for semi-synchronous scheduling. Without loss of generality suppose $r_{l1}$ is moving. $r_{l1}$ moves to a distance $2(rad_{req}-rad)$. Hence, the length of the diameter of this SEC is $2rad$, i.e., the radius is $2rad$. 
Under case 2, if both $r_{l2}$ and $r_{l2}$ move outward, after reaching the destination, the distance between $r_{li}$ and $r_{pi}$ $(i=1,2)$ is $2rad_c$. $r_{li}r_{pi}$ is not the diameter of the new SEC. However, $r_{li}r_{pi}$ is a chord of the new SEC. The actual diameter of the SEC is larger than $|r_{li}r_{pi}|= 2rad$. Hence, the diameter of the new SEC is $\ge 2rad$, i.e., the radius is $\ge 2rad$. 

\begin{lemma}
When there exists a robot at $p$ in case 1 or there exists a robot at $p_{i} (i=1,2)$ in case 2, the movement of $r_l$(case 1) or $r_{li} (i=1,2)$ (case 2) is collision free.
\end{lemma}
\textbf{Proof:}
Since the leaders are moving outside and no robot other than the leader is moving, we can state the lemma. Since, $r_l$ (case 1) or $r_{li} (i=1,2)$ (case 2) move diagonally outwards from the current SEC. No other robot is moving. hence no robot comes in the path of these moving robots.

\begin{lemma}
When there exists no robot at $p$ in case 1 or there exists no robot at $p_{i} (i=1,2)$ in case 2, and when $r_l$ (case 1) or $r_{li} (i=1,2)$ or $r'_{li} (i=1,2)$  (case 2) is moving to $q$ (case 1) or $q_i(i=1,2)$ (case 2), then $r_l$ (case 1) or $r_{li} (i=1,2)$ or $r'_{li} (i=1,2)$ (case 2) and $r_p$ (case 1) or $r_{pi} (i=1,2)$ (case 2) always remain on the current SEC.
\end{lemma} 
\textbf{Proof:} 
First consider case 1. Initially the distance between $r_l$ and $r_p$ is maximum. No robot other than $r_l$ moves. $|r_pq| > |r_pr_q|$. Hence, when $r_l$ reaches $q$ the distance between $r_p$ and $r_l$ remains maximum among other pairs of robots. According to the property of SEC the maximum distant points in a point set lie on the SEC of that point set. Hence, $r_l$ and $r_q$ lie on the new SEC. 
Now consider case 2. 
Suppose the situation when the path between $r_{li}$ and $q_{i}$ ((i=1 or 2) respectively) is free path. If any one of $r_{l1}$ or $r_{l2}$ moves, the case is similar to case 1. 
Otherwise, initially, $r_{pi} (i=1,2)$ lie at maximum distance from $r_{li} (i=1,2)$. Hence, $r_{pi} (i=1,2)$ is in maximum distance from $r_{li}(i=1,2)$. No robot other than $r_{li}(i=1,2)$  is moving. 
$|r_{pi}{qi}| > |r_{pi}r_{qi}| (i=1,2)$. Hence, when $r_{li}(i=1,2)$ reaches $q_i(i=1,2)$ the distance between $r_{pi}(i=1,2)$ and $r_{li}(i=1,2)$ remains maximum among other pairs of robots. According to the property of SEC the maximum distant points in a point set lie on the SEC of that point set. Hence, $r_{li}(i=1,2)$ and $r_{qi}(i=1,2)$ lie on the new SEC. 
Now consider the situation when the path between $r_{li}$ and $q_{i}$ ((i=1 and 2) respectively) is not free  path. 
Then $r'_{li}(i=1,2)$ moves to $q_i(i=1,2)$. Note that after the movement of $r'_{li}(i=1,2)$, the distance between $r'_{li}(i=1,2)$ and $r_{pi}(i=1,2)$ is maximum distance among any other pair of distances. Hence, according to the property of SEC  $r'_{li}(i=1,2)$ and $r_{pi}(i=1,2)$ lie on the new SEC. 

\begin{lemma}
\label{gerc}
When there exists no robot at $p$ in case 1 or there exists no robot at $p_{i} (i=1,2)$ in case 2, and when $r_l$ reaches $q$ (case 1), $r_{li}(i=1,2)$ or $r'_{li}$ reaches $q_{i}(i=1,2)$ (case 2), the radius of the new SEC $\ge rad_{req}$.  
\end{lemma}  
\textbf{Proof:} First consider case 1. The distance between $r_l$ at $q$ and $r_p$ is maximum among all pair distances. If $r_pq$ is the diameter of the new SEC then its radius is $= rad_{req}$. Otherwise, $r_pq$ is the chord of the new SEC where the actual diameter is $> 2rad$. Hence, the radius of the new SEC is $ >rad_{req}$.
Now consider case 2. Since $r_pi (i=1,2)$ and $q_i(i=1,2)$ are on the new SEC, with a similar argument as in case 1, it can be proved that the radius of the new SEC $\ge rad_{req}$.

\begin{lemma}
When there exists no robot at $p$ in case 1 or there exists no robot at $p_{i} (i=1,2)$ in case 2, the movement of $r_l$ (case 1) or $r_{li}$ (case 2) to $q$ (case 1) or $q_i(i=1,2)$ (case 2) is collision-free.	
\end{lemma}
\textbf{Proof:} 
Since the leaders are moving outside and no robot other than the leader is moving, we can state the lemma. First consider case 1. According to the algorithm $r_l$ moves only when there is a free path to $q$. Otherwise, the robot $r'_l$ having a free path to $q$ and nearest to $q$ moves to $q$. Thus there is no chance of collision as the robot moves along a free path. Case 2 can be proved  using similar arguments.

\begin{lemma}
If initially $rad < rad_{req}$, SECExpansion make $rad >= rad_{req}$ in a finite time.
\end{lemma}
\textbf{Proof:} The leader robots move to enlarge the SEC. Since the robots are semi-synchronous the leader does not change. Since the robots follow rigid motion, the leader successfully reaches its destination. Following lemmas \ref{gerc0} and \ref{gerc}, the radius of the new SEC is $\ge rad_{req}$.

\subsection{Description of The Algorithm ComputeTargetPoint}
The robots compute the target points on the SEC using ComputeTargetPoint. These are computed as equidistant points starting from the north-most intersection point of $L$ and SEC.\\
Let the north-most intersection point of $L$ and the SEC be $o$. $o$ is the first target point. If $rad$ is the current radius of SEC, then the next target point is computed as $\frac{2\pi rad}{n}$ distance apart from $o$ at both sides of $L$. Similarly all other target points are counted such that the distance between two consecutive target points is $\frac{2\pi rad}{n}$. Note that $rad \ge rad_{req}$.

\begin{algorithm}[H]
\KwIn{$n$, $rad$.}
\KwOut{$\{T_0, T_1, \ldots, T_n \}$: Equidistant target points on the circumference of the SEC.}
$o \leftarrow$ the north-most intersection point of $L$ and the SEC;
$T_0 \leftarrow o$;
$i=0$;\\
\While { $i \le {n-1}$}{
$T_{i+1}$ = A point on the circumference of the SEC, $\frac{2\pi rad}{n}$ unit apart from $T_i$;\\
i=i+1;}
return $\{T_0, T_1, \ldots, T_{n-1}\}$;
\caption{ComputeTargetPoint($n$, $rad$)}
\normalsize
\end{algorithm}

\subsection{Description of The Algorithm FormUCircle}
The autonomous unit-disc robots with unlimited visibility range, form the uniform circle by executing the algorithm FormUCircle as described below.

\begin{definition}
A vacant target position is a point on the circumference of the SEC such that there exist no parts of another robot around a circular region of two units radius around this point.
\end{definition}

\begin{itemize}
\item The robots which lie on the circumference of the circle, they slide along the perimeter of the circle to their destinations. The robots lying inside the circle, move in a straight line to their destinations.     
\item Let $T$ be a vacant target point having maximum Y value (north most). In case of symmetry there may be two such points. Both points will be considered in the same priority. 
\item A robot $r$ will move to a target point obeying the following strategy. 
\begin{itemize}
\item If $r$ is nearest to a north-most target point $T$, $r$ moves to $T$. 
If there are multiple robots nearest to $T$, the robot having a maximum Y value among them moves to $T$. However, if $T=o$, then no robot move. 
\item If $r$ is nearest to more than one north-most target points with no competent robots, then any one of the target points is chosen arbitrarily.
\end{itemize}
\end{itemize}

If there exists any robot  $r_o$ in the path of a robot $r_i$ towards its target point $T$, the $r_i$ slides over $r_o$ and moves to $T$. Note that if there are multiple robots at the same distance from a target then the north most robot is selected for movement. Hence this robot can never be obstructed by both sides to move to its destination. 

Observe that it is possible for a robot $r$ which is already in a target point $T1$, to move to another target point $T2$, since, $R$ is nearest to $T2$ and there exists no other robot which may move to $T2$. However, eventually, this shifting from the target point phenomenon will stop after a finite number of execution cycles.

\begin{algorithm}[H]
\KwIn{$n$.}
\KwOut{A robot $R$ reaches its target point on the SEC.}
Compute the SEC of $n$ robots;\\
$rad \leftarrow$ radius of the SEC;
$rad_{req} \leftarrow ComputeRadius(R)$\\
\eIf{if $rad \le rad_{req}$}
{SECExpansion(R)}
{ComputeTargetPoint(R);\\ 
$\cal{T} \leftarrow$ the north most target point nearest to $R$\\
\eIf{${\cal{T}} = T_0$ and there exists another robot $R'$ nearest to $T_0$}{$R$ does not move}
{\eIf{there exist multiple robots  $\{r_1, \ldots, r_k\}$ nearest to $\cal T$}
{ $r_x \leftarrow$ a robot in $\{r_1, \ldots, r_k\}$ with maximum $Y$ value\\
\eIf{$R = r_x$}
{$R$ moves to $\cal T$\;}{$R$ does not move;}
} 
{ \eIf{$R$ is nearest to multiple target points $\{{\cal T}_1, \ldots, {\cal T}_p\}$}
{$\cal{T'} \leftarrow$ a target point in $\{{\cal T}_1, \ldots, {\cal T}_p\}$ with maximum $Y$ value;\\
$R$ moves to $\cal{T'}$;}
{$R$ moves to $\cal{T}$;
}
}
}
}
\caption{FormUCircle(n)}
\normalsize
\end{algorithm}

\begin{lemma}
FormUcircle and SECExpansion will not be executed simultaneously.
\end{lemma}
\textbf{Proof:}
The robots will execute FormUcircle only when $rad \ge rad_{req}$. The robots will execute SECExpansion only when $rad < rad_{req}$. Both predicates can not be true simultaneously. Hence, both algorithms will not be executed together.

\begin{lemma}
When $R$ is moving to $T$ if any other robot computes (due to asynchrony), $R$ remains the only candidate eligible to move to $R$.
\end{lemma}
\textbf{Proof:}
Since $R$ is nearest to $T$, when it is moving towards $T$, it becomes more closer to $T$. Thus if any other robot computes, it finds $R$ as nearest to $T$. Hence there is no chance for another robot to move to $T$.

\begin{lemma}
When a robot $r_T$ is moving to $T$, no other robot comes in its path, i.e., the movement of $r_T$ is collision-free.	
\end{lemma}
\textbf{Proof:} 
In this algorithm, the robot nearest to the vacant north-most target point, moves to the target point or existence of any obstacle robot following the below two situations may arise.
\begin{itemize}
\item The obstacle robot is nearer to the target point, which is not possible.
\item The moving robots can be obstructed by both sides when three robots are at the same distance from the target and the middle robot touches the other two robots from both sides. 
In this situation the north-most robot is selected from the movement. This robot will have an open side and it will slide over the other robot and moves to its destination.
\end{itemize}
Thus the robots reach their destinations without collision.

Through our algorithm, all the robots will have their target points and path to travel to reach it. If no target point is vacant; $n$ target points are partially or fully filled by $n$ robots; Then all the robots will move to those target points occupied by them partially. Otherwise there exists at least one vacant target point. Note that since the side of the polygon is $>3$ units, no robot can partially block two target points.  
Hence, the following lemma holds.

\begin{lemma}
There will be no deadlock in the formation of regular polygon.	
\end{lemma}
\textbf{Proof:}  
The vacant target points from the north-most side are getting filled by the robots. Since the number of target points is equal to the number of robots there exists always a vacant target point to be filled by its nearest robot. 
In ordering of the robots' movement is maintained implicitly in this algorithm, thus there is no deadlock. 


\section{Uniform Circle Formation By Swarm Robots Under Limited Visibility}
In this section, we first describe the robot model used in this paper and present an overview of the problem. Then we move to the solution approach and present the algorithms with the proofs of their correctness.
\subsection{Underlying Model}
We use the basic structure of the weak model \cite{Ref5} of robots with some extra features which extend the model towards the real-time situation. Let R = {$r_{1}, r_{2},..,r_{n}$}
be a set of unit disc-shaped autonomous robots referred to as {\it fat robots}. A robot is represented by its center i.e., by $r_{i}$ we mean a robot whose center is $r_{i}$. The set of robots R deployed on the 2D plane is described as follows:
\begin{itemize}
    \item The robots are autonomous.
    \item Robots are anonymous and homogeneous in the sense that they are unable to uniquely identify themselves, neither with a unique identification number nor with some external distinctive mark (e.g. color, flag). 
    \item The robots are oblivious in the sense that they can not recollect any data from the past cycle.
    \item The robots have rigid movement.
    \item Robots can not communicate explicitly. Each robot is allowed to have a camera that can take pictures over 360 degrees and up to a fixed radius. The robots communicate only by means of observing other robots within its visibility range with the camera.
    \item The robots are non-transparent, i.e., opaque and also act as obstacle for other robots.
    \item A robot can see up to a fixed distance around itself on the 2D plane.
    \item Each robot executes a cycle of wait-look-compute-move asynchronously.
     \item The robots have a common origin, common x-y axis, a common sense of direction and common unit distance.
\end{itemize}
\subsection{Overview of the Problem}
A set of robots R (as described above) is given. Our objective is to form a circle (denoted by CIR) of radius rad and centered at C by moving the robots from R.
Following assumptions and definitions are used in this paper:

\begin{definition} Each robot can see up to a fixed distance around itself. This distance is called the visibility range of that robot. The visibility range of $r_{i} \in$ R is denoted by $R_{v}$ and is equal for all robots in R.
\end{definition}

\begin{figure}[H]
    \centering
   \includegraphics[scale=0.59]{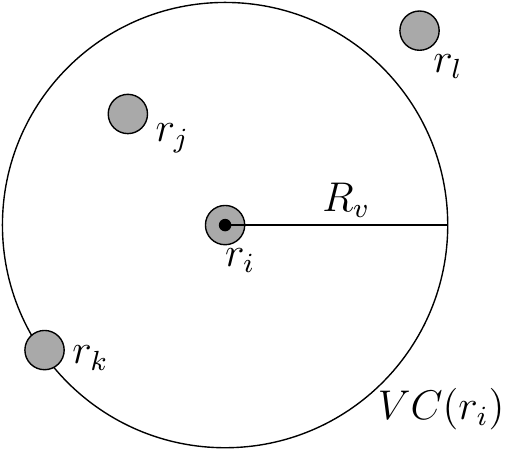} 
   \caption{Visibility Radius $(R_{v})$ and Visibility Circle $(VC(r_{i}))$}
   \label{ii_1}
\end{figure}

\begin{definition}
The circle, centered at robot $r_{i}$ and having radius $R_{v}$ is called the visibility circle of the robot $r_{i}$, denoted by VC($r_{i}$). $r_{i}$ can see everything within and on the circumference of VC($r_{i}$), but cannot see beyond VC($r_{i}$) (Figure \ref{ii_1})
\end{definition}

\textbf{Assumptions:}
\begin{itemize}
\item All robots in R agree on a common origin, axes, sense of direction and unit distance. C, the center of the circle CIR is considered as the origin of the coordinate system.
\item The radius (rad)of the circle to be formed (CIR) is given. The length of rad is such that CIR can accommodate all the robots in R.
\item Initially the robots in R can be either inside, outside or on the CIR.
\end{itemize}

\paragraph{\textbf{Notations:}} The following notations are used throughout the paper:
\begin{itemize}
\item CIR: Circle to be formed, with radius, rad, and centered at C.
\item T($r_{i}$): Destination point for robot $r_{i}$.
\item rad: The given radius of the circle to be formed.
\item dist(p1, p2): Euclidean distance between two points p1 and p2.
\item cir($r_{i}$, C): A circle centered at C and having a radius of dist($r_{i}$,C).
\item projpt($r_{i}$, A): The projected (radially outward) point of the robot position $r_{i}$ on circle A.
\item arc(a, b): The arc of a circle between the points a and b on the circumference of that circle.
\end{itemize}


Two constraints have been put on the movement of any robot $r_{i}$.

\begin{figure}[H]
      \centering
   \includegraphics[scale=0.61]{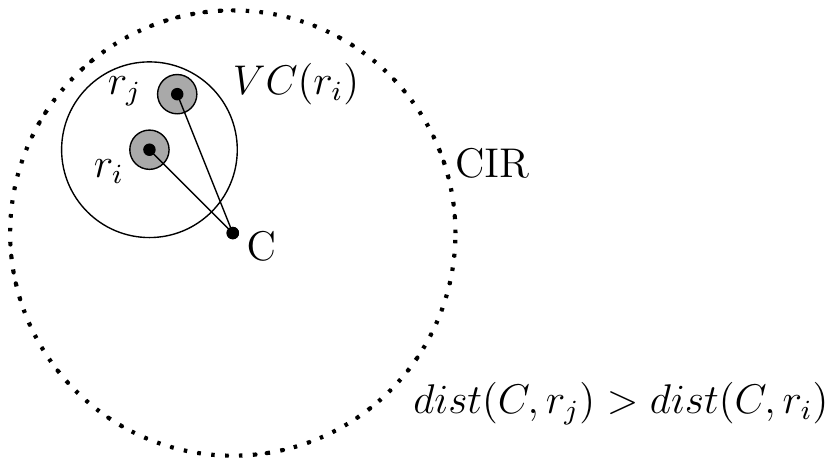} 
   \caption{An example to represent constraint 1}
   \label{ii_2}
\end{figure}

\textbf{Constraint 1: }
Let $r_{j} \in$ R be any robot inside VC($r_{i}$) (the visibility circle of $r_{i}$). $r_{i}$ is eligible to move if----
\begin{itemize}
\item $r_{i}$ is inside CIR, dist(C; $r_{i}$) $\geq$ dist(C; $r_{j}$) and dist(C; $r_{j}$) $<$ rad.
\item $r_{i}$ is inside CIR, and dist(C; $r_{i}$) $<$ dist(C; $r_{j}$) and dist(C; $r_{j}$) = rad.
\item $r_{i}$ is outside CIR, dist(C; $r_{i}$) $\leq$ dist(C; $r_{j}$) and dist(C; $r_{j}$) $>$ rad.
\item If $r_{i}$ is outside CIR, and dist(C; $r_{i}$) $>$ dist(C; $r_{j}$) and dist(C; $r_{j}$) = rad.
\item If $r_{i}$ is at C, then $r_{i}$ is eligible to move.
\end{itemize}
Note: For all other cases $r_{i}$ will not move. (Figure \ref{ii_2})

\textbf{Constraint 2: }
$r_{i} \in$ R moves only in any of the following fixed directions.
\begin{itemize}
\item Radially outwards following the ray starting from C, directed towards $r_{i}$.
\item Radially inwards following the ray starting from $r_{i}$, directed towards C.
\item Right side of the ray, starting from C and directed towards $r_{i}$.
\item Right side of the ray, starting from $r_{i}$ and directed towards C.
\end{itemize}

This following sections present the description of algorithms ComputeTargetPoint(n,rad), ComputeRobotPosition(n,rad), ComputeDestination(R) and UniformCircleFormation(R) required for uniform circle formation by computing the destination for the robots. 
\begin{itemize}
\item First the target positions of the robots are computed in {\it ComputeTargetPoint}.
\item Then the current positions of the robots with respect to the CIR are decided in {\it ComputeRobotPosition}.
\item Next the destination positions of the robots on the CIR are computed in {\it ComputeDestination}.
\item Finally the uniform circle is formed by executing the {\it UniformCircleFormation} algorithm.
\end{itemize}

\subsection{Description of The Algorithm ComputeTargetPoint}
Let $L$ be the line parallel to the Y-axis and passing through the center $C$ of CIR. If the north-most intersection point of $L$ and CIR be $o$ then, $o$ is the first target point. The next target point is computed as $\frac{2\pi rad}{n}$ distance apart from $o$ at both sides of $L$. Similarly all other target points are counted such that the distance between two consecutive target points is $\frac{2\pi rad}{n}$. The target points are computed by the ComputeTargetPoint($n$, $rad$) algorithm (Refer to {\it section 3.4}).


\subsection{Description of The Algorithm ComputeRobotPosition}
This algorithm decides the position of robot $r_i$, either inside CIR or outside CIR or on the CIR.
The inputs to the algorithm are $n$ and $rad$.
Let coordinates of the center of CIR, C is (0,0) and the position of $r_i$ is (x,y).
 For all $r_i$, if $\sqrt{(0-x)^2+(0-y)^2} = CIR$, then $r_i$ is on the CIR. Else if $\sqrt{(0-x)^2+(0-y)^2} < CIR$, then $r_i$ is inside CIR. Otherwise, $r_i$ is outside CIR.

\begin{algorithm}[H]
\caption{ComputeRobotPosition($n$, $rad$)}
\KwIn{$n$, $rad$.}
\KwOut {Position of robot $r_i$; either inside CIR or outside CIR}
Center of CIR, C $\leftarrow$ (0,0) and position of $r_i$ $\leftarrow$ (x,y)\\
For all $r_i$,
If $\sqrt{(0-x)^2+(0-y)^2} < CIR$ then $r_i$ is inside CIR\\
Else $r_i$ is outside CIR.\\
Return Set $r_i$ that are inside CIR and $r_i$ that are outside CIR;
\end{algorithm}

\subsection{Description of The Algorithm ComputeDestination}
\begin{figure}[H]
            \centering
            \includegraphics[scale=0.51]{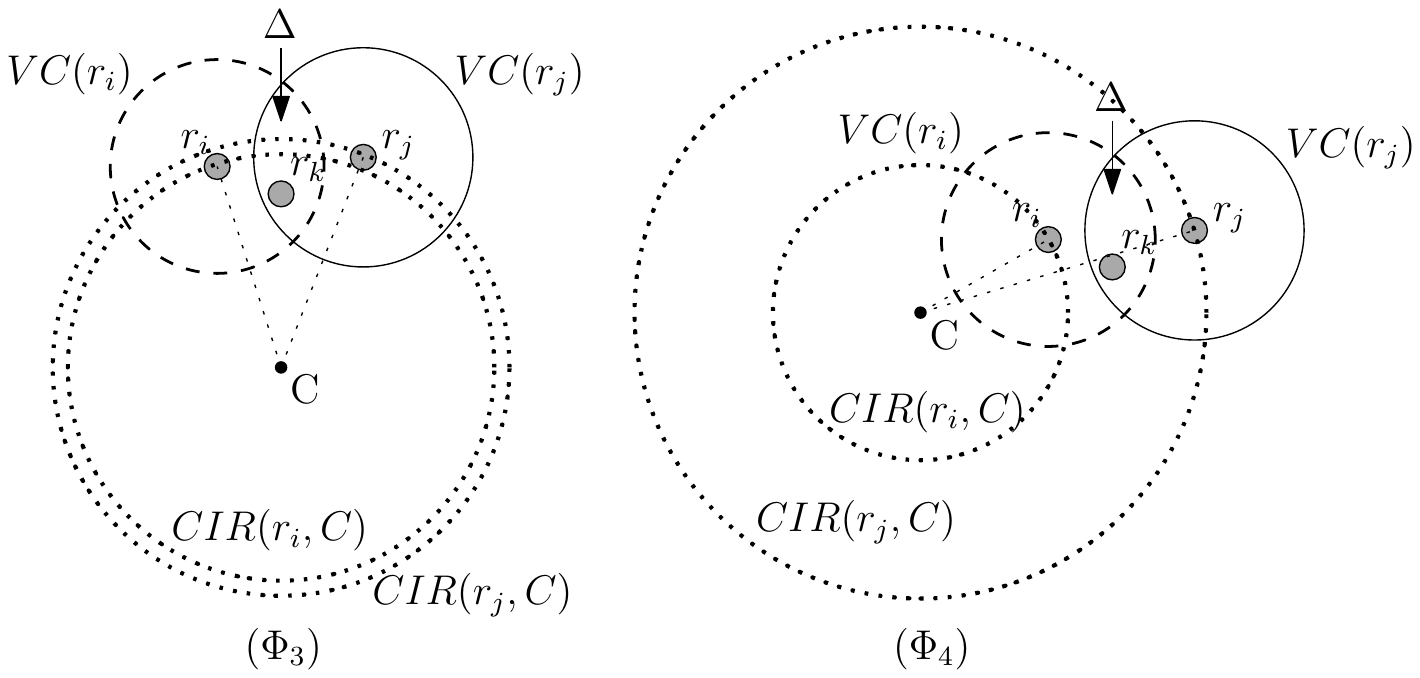} 
            \caption{An example of the configurations $\Phi$3 and $\Phi$4}
            \label{ii_5}
        \end{figure}

We categorize different configurations depending on the position of visibility circles of $r_{i}$ and $r_{j}$ . We denote these configurations as $\Phi$1, $\Phi$1, $\Phi$3 and $\Phi$4.
\begin{itemize}
    \item $\Phi$1: \textbf{V($r_{i}$) and V($r_{j}$) do not touch or intersect each other (Figure \ref{ii_3}).}
    
    \begin{figure}[H]
            \centering
            \includegraphics[scale=0.35]{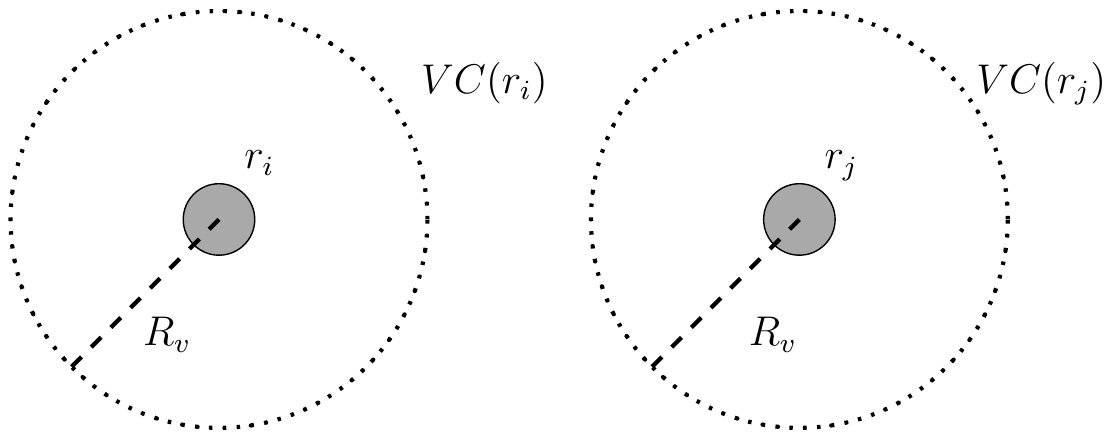} 
            \caption{An example of the configuration $\Phi$1}
            \label{ii_3}
        \end{figure}
        
    \item $\Phi$2: \textbf{V($r_{i}$) and V($r_{j}$) touch each other at a single point (say k) (Figure \ref{ii_4}).} If there is a robot at k say $r_{k}$, then $r_{i}$ and $r_{k}$ and $r_{j}$ and $r_{k}$ are mutually visible.
    
     \begin{figure}[H]
            \centering
            \includegraphics[scale=0.51]{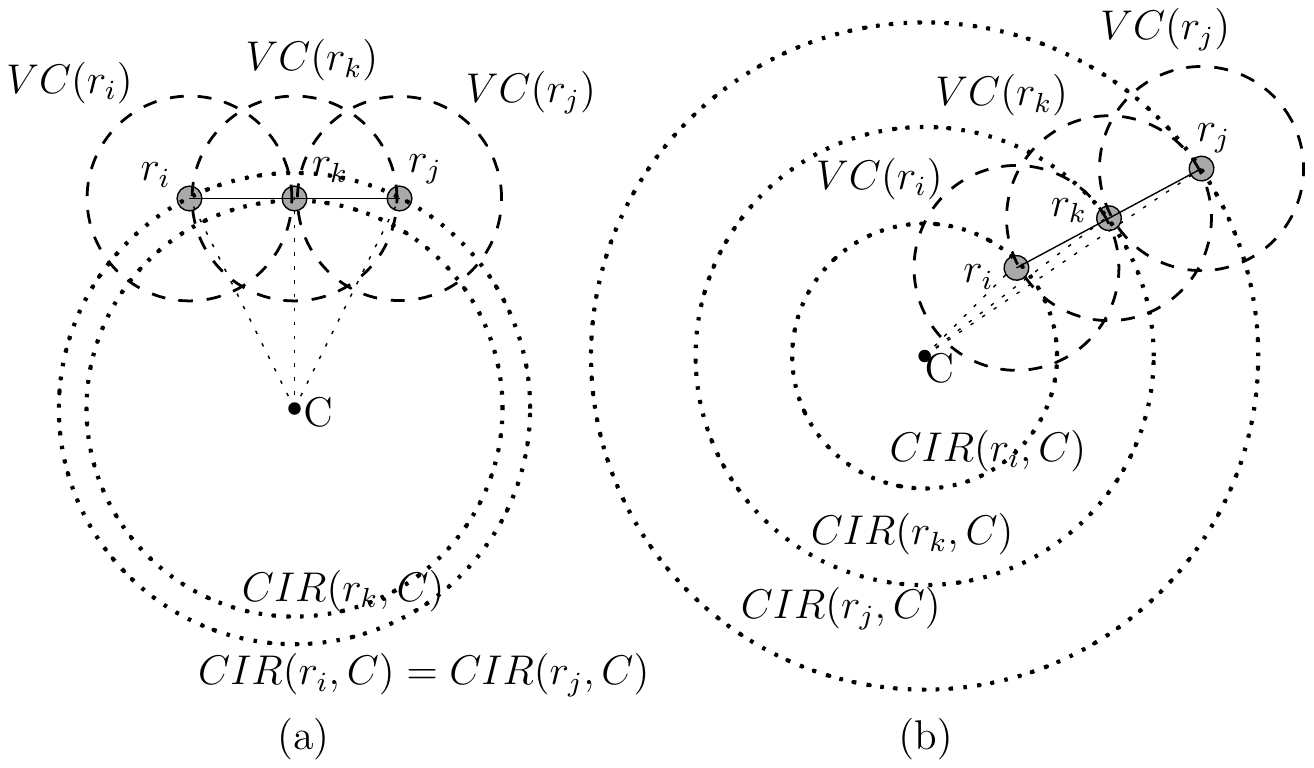} 
            \caption{An example of the configuration $\Phi$2}
            \label{ii_4}
            \end{figure}
        
    \item $\Phi$3: \textbf{V($r_{i}$) and V($r_{j}$) intersect each other at two points such that $r_{i}$ and $r_{j}$ can not see each other (Figure \ref{ii_5} $\Phi$3).} Let $\Delta$ be the common visible region of $r_{i}$ and $r_{j}$. If there is a robot in the region $\Delta$, say $r_{k}$, then $r_{k}$ can see $r_{i}$ and $r_{j}$ , and both $r_{i}$ and $r_{j}$ can see $r_{k}$.

    \item $\Phi$4: \textbf{V($r_{i}$) and V($r_{j}$) intersect each other at two points such that $r_{i}$ and $r_{j}$ can see each other (Figure \ref{ii_5} $\Phi$4).} Let $\Delta$ be the common visible region of $r_{i}$ and $r_{j}$ . If there is a robot in $\Delta$ region, say $r_{k}$, then $r_{k}$, $r_{i}$ and $r_{j}$ can see each other.
      
\end{itemize}
\vspace{5mm}        
There are ten configurations depending on the position of $r_{i}$ inside or outside CIR. We denote these configurations as 	$\Psi_{0}$, $\Psi_{1}$, $\Psi_{2}$, $\Psi_{3}$, $\Psi_{4}$, $\Psi_{5}$, $\Psi_{6}$, $\Psi_{7}$, $\Psi_{8}$ and $\Psi_{9}$.

\begin{itemize}
    \item $\Psi_{0}$: \textbf{$r_{i}$ is on the CIR circumference and vacant space is available, radially outside CIR.}  $r_{i}$ moves radially outward to the available vacant space, else $r_{i}$ does not move until vacant space is available outside CIR.
    \item $\Psi_{1}$: \textbf{$r_{i}$ is on a target point, on the circumference of CIR, it does not move any further (Figure \ref{ii_6}).}
    
    \begin{figure}[H]
        \centering
        \includegraphics[scale=0.46]{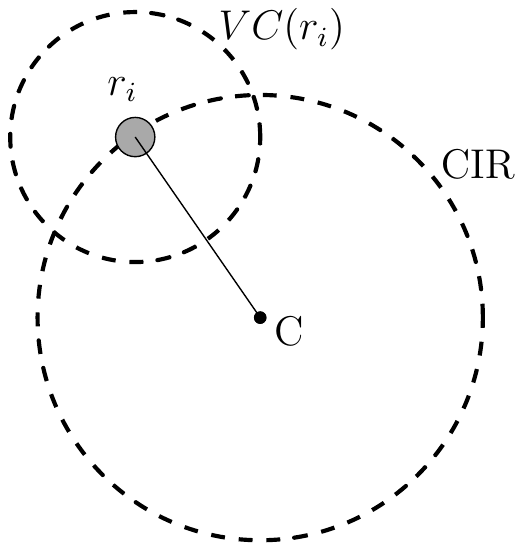} 
        \caption{An example of the configuration $\Psi_{1}$}
        \label{ii_6}
    \end{figure}
  
    \item $\Psi_{2}$: \textbf{$r_{i}$ is inside the CIR and VC($r_{i}$) touches CIR (at a point, say h) (Figure \ref{ii_7}).} If h is vacant and is a target position, then $T(r_{i})$ moves to h. Otherwise, $r_{i}$ moves to the midpoint of the line joining $r_{i}$ and h.
    
    \begin{figure}[H]
        \centering
        \includegraphics[scale=0.49]{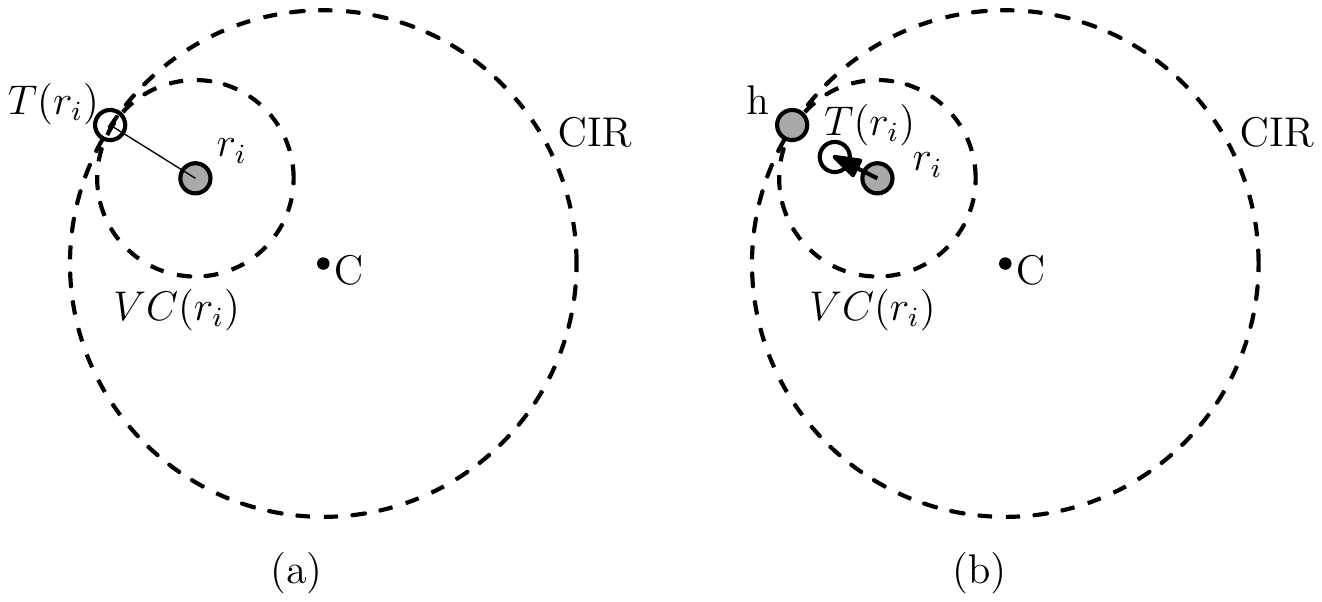} 
        \caption{An example of the configuration $\Psi_{2}$}
        \label{ii_7}
    \end{figure}
    
    \item $\Psi_{3}$: \textbf{$r_{i}$ is inside the CIR but not at C and VC($r_{i}$) does not touch or intersect the circumference of CIR (Figure \ref{ii_8}).}
    Let t be  projpt($r_{i}, VC(r_{i})$, If t is a vacant point, then $r_{i}$ moves to t. Otherwise, $r_{i}$ moves to the midpoint of the line joining $r_{i}$ and t.

    \begin{figure}[H]
        \centering
        \includegraphics[scale=0.49]{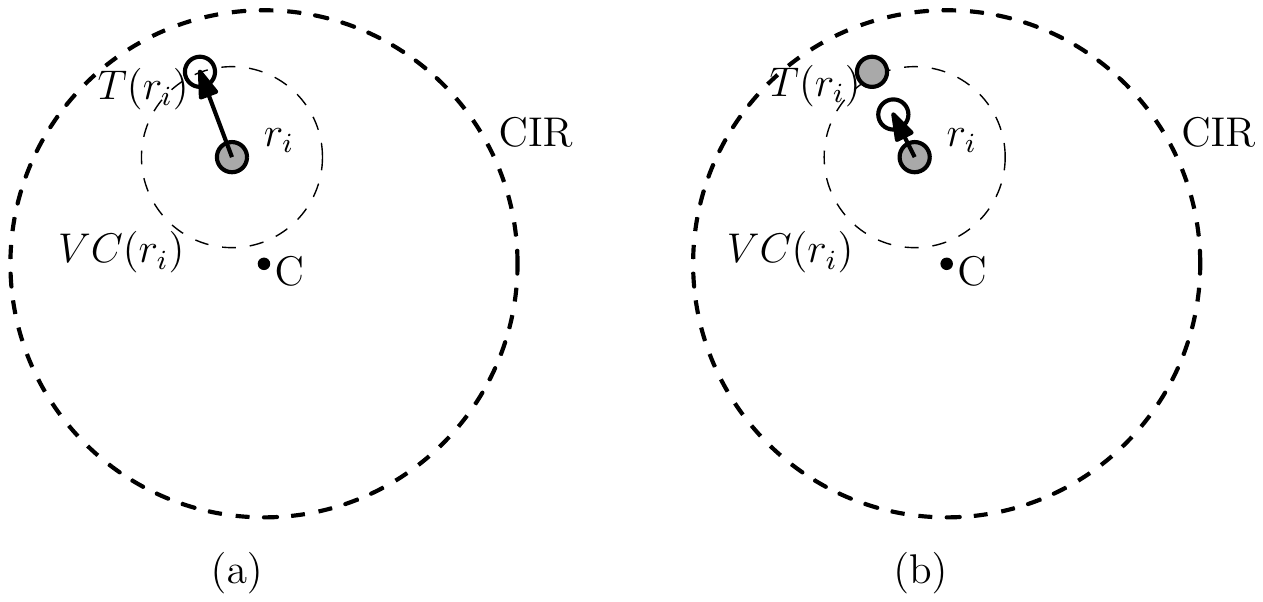} 
        \caption{An example of the configuration $\Psi_{3}$}
        \label{ii_8}
    \end{figure}

    \item $\Psi_{4}$: \textbf{$r_{i}$ is at center, C (Figure \ref{ii_9}).} It moves to the intersection point of positive X-axis of robot $r_{i}$ and $VC(r_{i})$ (Say m). If m is vacant then $T(r_{i})$ moves to m, else $T(r_{i})$ moves to the midpoint of the line joining $r_{i}$ and m.
   
    \begin{figure}[H]
        \centering
        \includegraphics[scale=0.52]{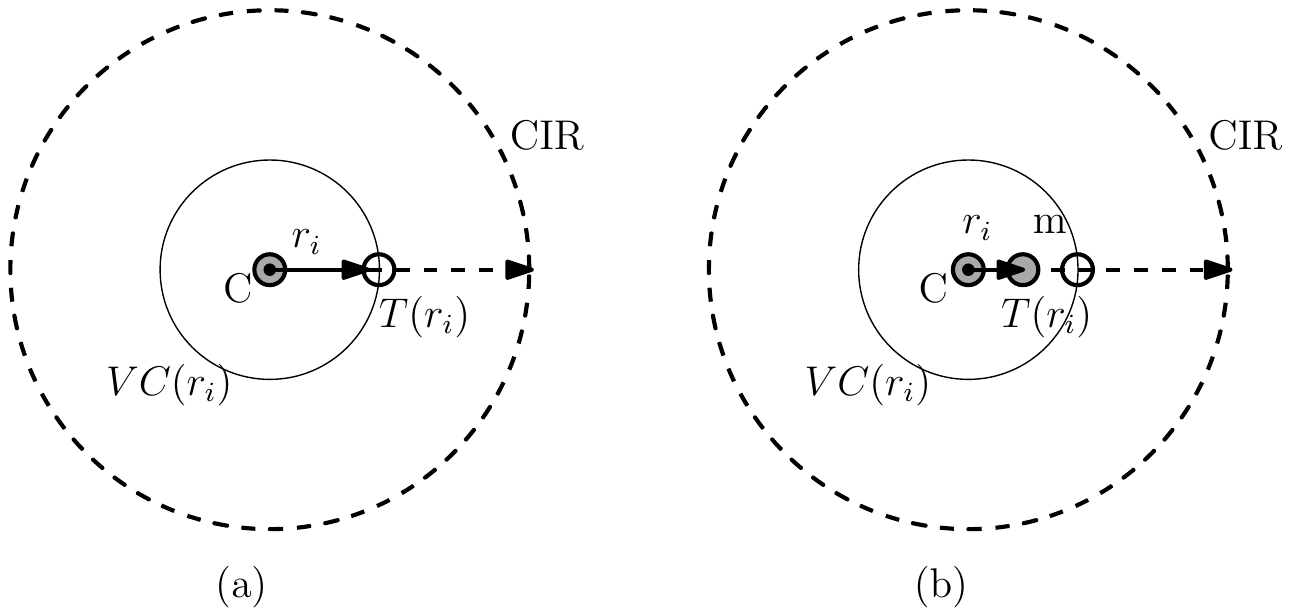} 
        \caption{An example of the configuration $\Psi_{4}$}
        \label{ii_9}
    \end{figure}
        
    \item $\Psi_{5}$: \textbf{$r_{i}$ is inside the CIR and VC($r_{i}$) intersect CIR (at two points say g and l) (Figure \ref{ii_10}).} Here, we can visualize the configuration of robots as m concentric circles whose center is C. We consider that the robots of $\Psi_{5}$ are in concentric circle C(m-1) and they can either jump to C(m)[as in Case $\Psi_{5}$(a) and Case $\Psi_{5}$(b)] or remain in the same circle C(m-1) to move rightwards [as in this case $\Psi_{5}$(c)].
	Possible cases for $\Psi_{5}$:
	\textbf{$\Psi_{5}$(a) - There is a target position $T(r_{i})$ on the radially outward projpt(r,C) of the arc gl.}
	\begin{itemize}
	    \item If $T(r_{i})$ is vacant then $r_{i}$ moves to $T(r_{i})$.
	    \item Else Check for next target point on the right of it upto l.
	\end{itemize}
	
	\textbf{$\Psi_{5}$(b) - There is no target point on the radially outward projpt(r,C) but there is a target position on the arc gl.}
	\begin{itemize}
	    \item If $T(r_{i})$ is vacant then $r_{i}$ moves to $T(r_{i})$.
	    \item Else Check for next target point on the right of it up to l.
	\end{itemize}
	
	\textbf{$\Psi_{5}$(c) -  There is no vacant target position at all on the arc tl.}
	\begin{itemize}
	    \item If the next position on the right on the same concentric circle C(m-1) is vacant, the robot on C(m-1) moves rightwards to the next position.
	        \begin{itemize}
	            \item If T(ri) is found on C(m) then move to T(ri) and exit $\Psi_{5}$;
	            \item If T(ri) not found then repeat $\Psi_{5}$(c).
	        \end{itemize} 
	    \item If the next position on the right on the same concentric circle C(m-1) is not vacant:
	    The robot, $r_{i}i$ on C(m-1) does not move until the robot on the next right position of $r_{i}$, has moved to C(m).
	\end{itemize}

    \begin{figure}[H]
        \centering
        \includegraphics[scale=0.53]{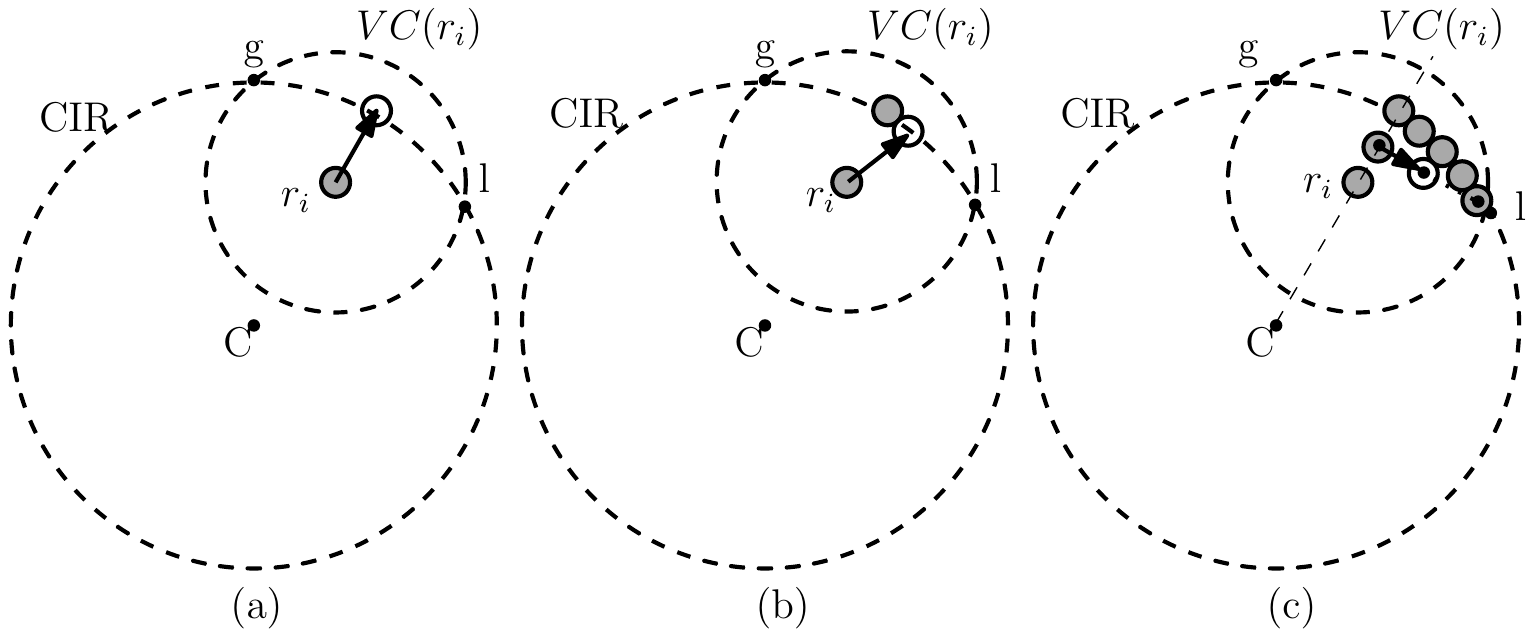} 
        \caption{An example of the configuration $\Psi_{5}$}
        \label{ii_10}
    \end{figure}
    
   \item $\Psi_{6}$: \textbf{$r_{i}$ is outside the CIR and VC($r_{i}$) touches CIR (at some point say h) (Figure \ref{ii_11}).} If h is a vacant point, then $r_{i}$ moves to h. Otherwise, $r_{i}$ moves to the midpoint of the line joining $r_{i}$ and h.
   
   \begin{figure}[H]
        \centering
        \includegraphics[scale=0.53]{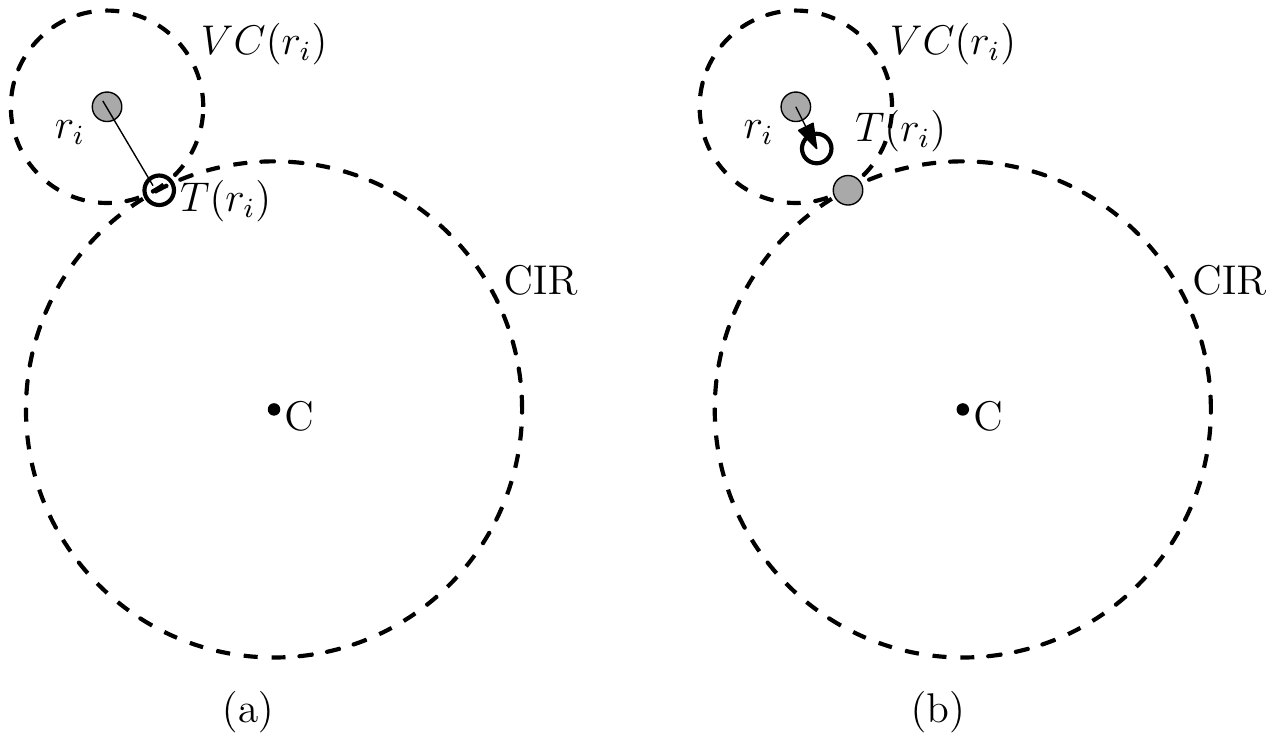} 
        \caption{An example of the configuration $\Psi_{6}$}
        \label{ii_11}
    \end{figure}
  
    \item $\Psi_{7}$: \textbf{$r_{i}$ is outside the CIR and VC($r_{i}$) do not touch or intersect the circumference of CIR (Figure \ref{ii_12}).} Let, t be projpt($r_{i}$, $VC(r_{i})$). If t is a vacant point, then $r_{i}$ moves to t. Otherwise, $r_{i}$ moves to the midpoint of the line joining $r_{i}$ and t.
    
    \begin{figure}[H]
        \centering
        \includegraphics[scale=0.53]{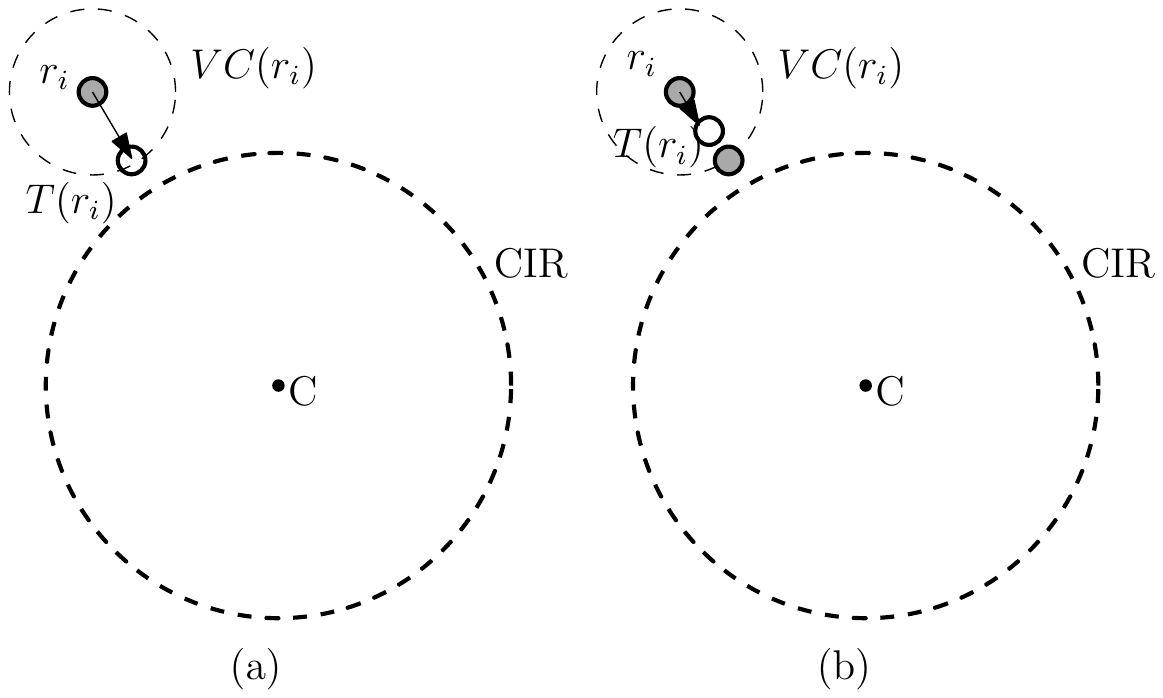} 
        \caption{An example of the configuration $\Psi_{7}$}
        \label{ii_12}
    \end{figure}

    \item $\Psi_{8}$: \textbf{$r_{i}$ is outside the CIR and VC($r_{i}$) intersect CIR (at two points say g and l) (Figure \ref{ii_13}).} Here, we can visualize the configuration of robots as m concentric circles whose center is C. We consider that the robots of $\Psi_{8}$ are in concentric circle C(m+1) and they can either jump down to C(m)[as in Case $\Psi_{8}$(a) and Case $\Psi_{8}$(b)] or remain in the same circle C(m+1) to move rightwards [as in this case $\Psi_{8}$(c)].Possible cases for $Psi_{8}$:
	\textbf{$\Psi_{8}$(a) - There is a target position $T(r_{i})$ on the radially inward projpt(r,C) of the arc gl.}
	\begin{itemize}
	    \item If $T(r_{i})$ is vacant, $r_{i}$ moves to $T(r_{i})$.
	    \item Else Check for next target point on the right of it up to l.
	\end{itemize}
	
	\textbf{$\Psi_{8}$(b) - There is no target point on the radially inward projpt(r,C) but there is target position on the arc gl.}
	\begin{itemize}
	    \item If $T(r_{i})$ is vacant, $r_{i}$ moves to $T(r_{i})$.
	    \item Else Check for next target point on the right of it up to l.
	\end{itemize}
	
	\textbf{$\Psi_{8}$(c) - There is no vacant target position at all on the arc tl.}
	\begin{itemize}
	    \item If the next position on the right on the same concentric circle C(m+1) is vacant, the robot on C(m+1) moves rightwards to the next position.
	        \begin{itemize}
	            \item If T(ri) found on C(m), move to T(ri) and exit $\Psi_{8}$;
	            \item If T(ri) not found, repeat $\Psi_{8}$(c). ;
	        \end{itemize} 
	    \item If the next position on the right on the same concentric circle C(m+1) is not vacant:
	    The robot, $r_{i}i$ on C(m+1) does not move until the robot on the next right position of $r_{i}$, has moved down to C(m).
	 \textbf{$\Psi_{8}$(c) -  There is no vacant target position at all on the arc tl.}  
	\end{itemize}
 \begin{figure}[H]
        \centering
        \includegraphics[scale=0.53]{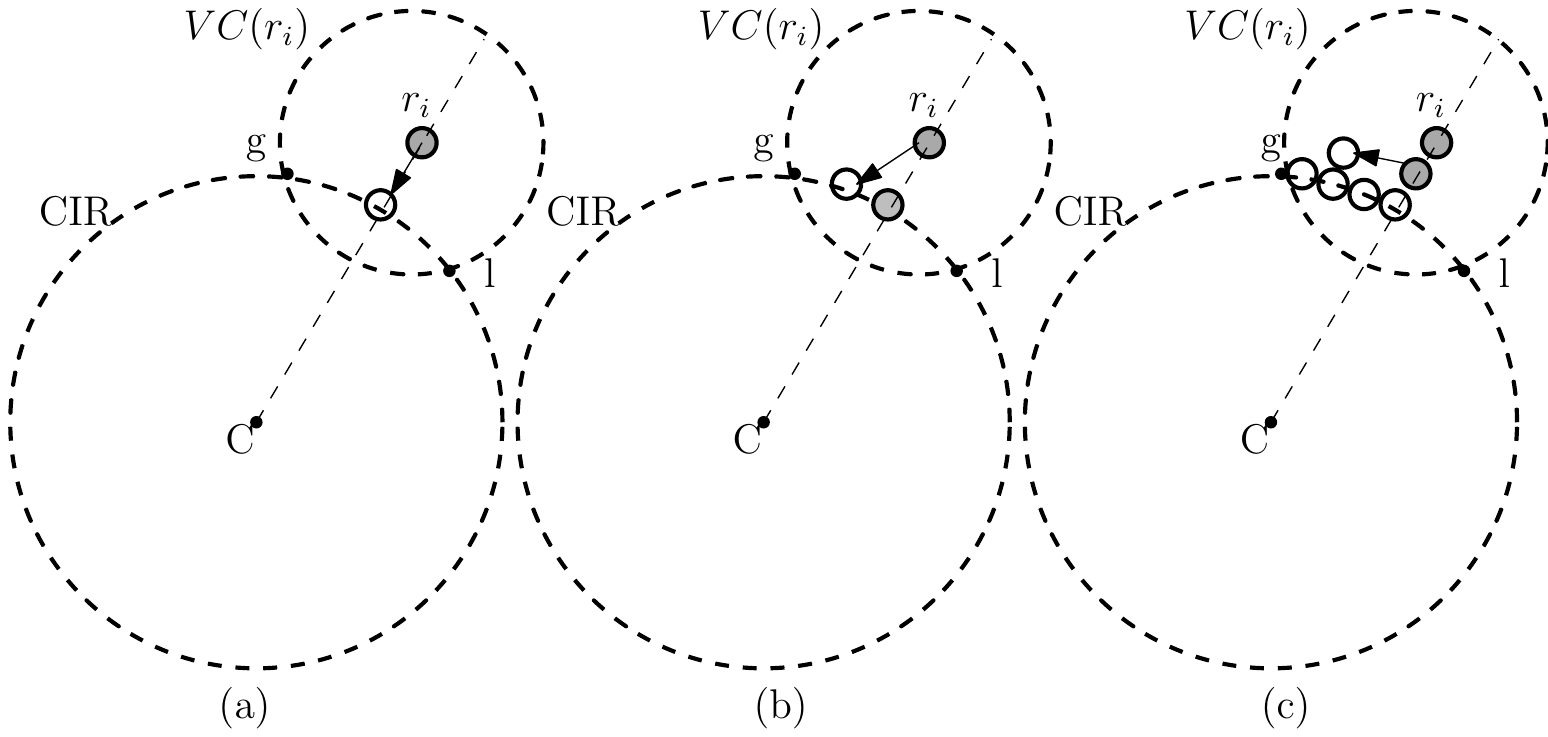} 
        \caption{An example of the configuration $\Psi_{8}$}
        \label{ii_13}
    \end{figure}

     \item $\Psi_{9}$: 
    \textbf{(i) - $r_{i}$ is outside the CIR with VC($r_{i}$) touching CIR at a point and there is a target position $T$ at that point. Also, another robot $r_{j}$ is inside the CIR with VC($r_{j}$) touching CIR at the same point. (Figure \ref{ii_14}).}
    \begin{itemize}
	    \item The robots $r_{i}$ moves radially inwards and $r_{j}$ moves radially outwards towards the CIR till the target point $T$ is visible to both $r_{i}$ and $r_{j}$.
	    \item The robot inside the CIR $r_{j}$ will move to the target position $T$ and the robot outside CIR, $r_{i}$ will move to the next vacant target position on its right.
	\end{itemize}

    \begin{figure}[H]
        \centering
        \includegraphics[scale=0.54]{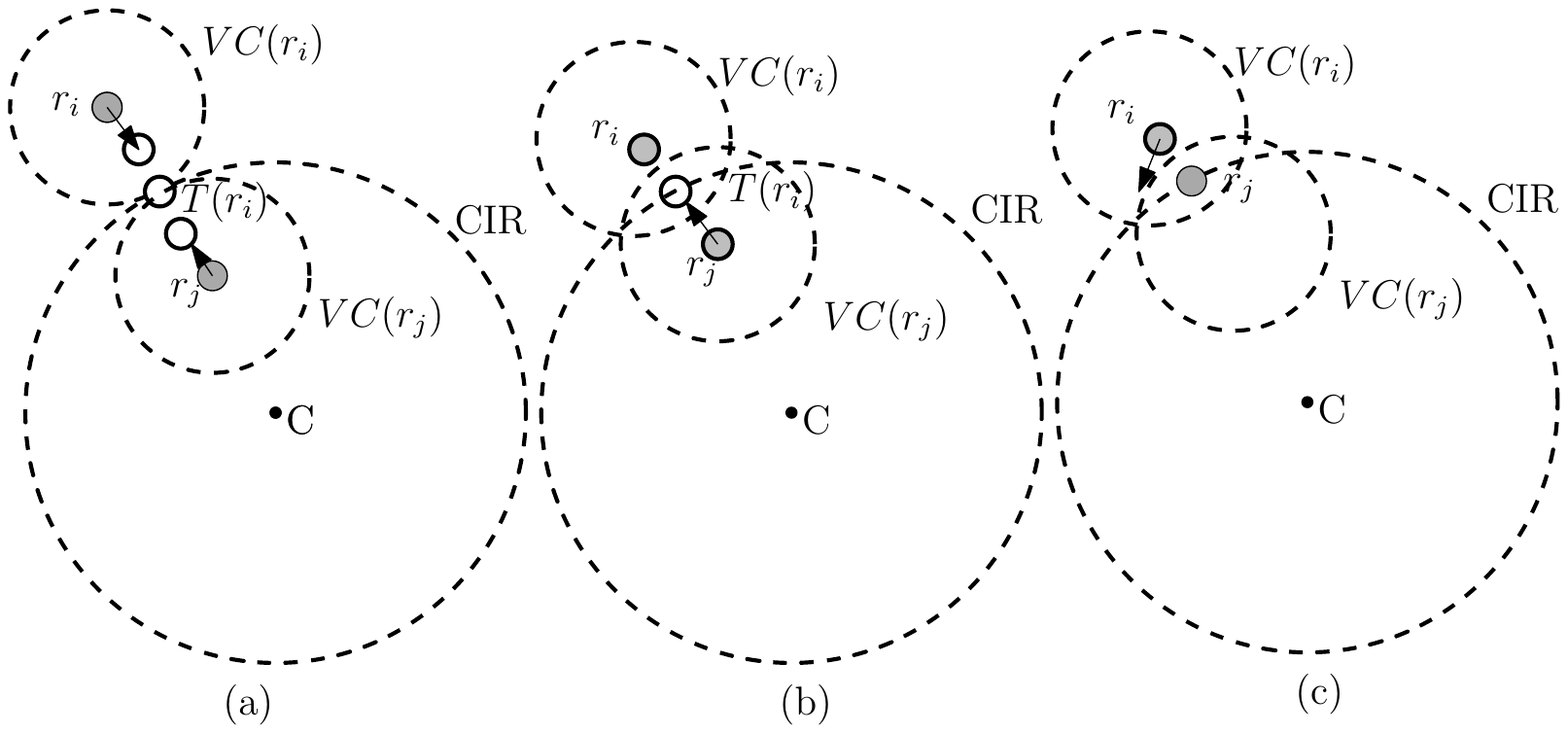} 
        \caption{An example of the configuration $\Psi_{9}$(i)}
        \label{ii_14}
    \end{figure}
    
    \textbf{(ii) - $r_{i}$ is outside the CIR with VC($r_{i}$) intersecting CIR (at two points say g and l) and $r_{j}$ is inside the CIR with VC($r_{j}$) intersecting CIR (Figure \ref{ii_15}).}
    \begin{itemize}
	    \item The robot inside CIR, $r_{j}$ will move radially outwards towards the CIR and occupy the vacant target position $T$, on the CIR.
	    If $T(r_{i})$ is not vacant $r_{j}$ moves as in configuration $\Psi_{8}$(c).
	    \item The robot outside the CIR, $r_{i}$ will move to the next available vacant target position on its right side, on the CIR as in configuration $\Psi_{8}$(c).
	\end{itemize}

    \begin{figure}[H]
        \centering
        \includegraphics[scale=0.58]{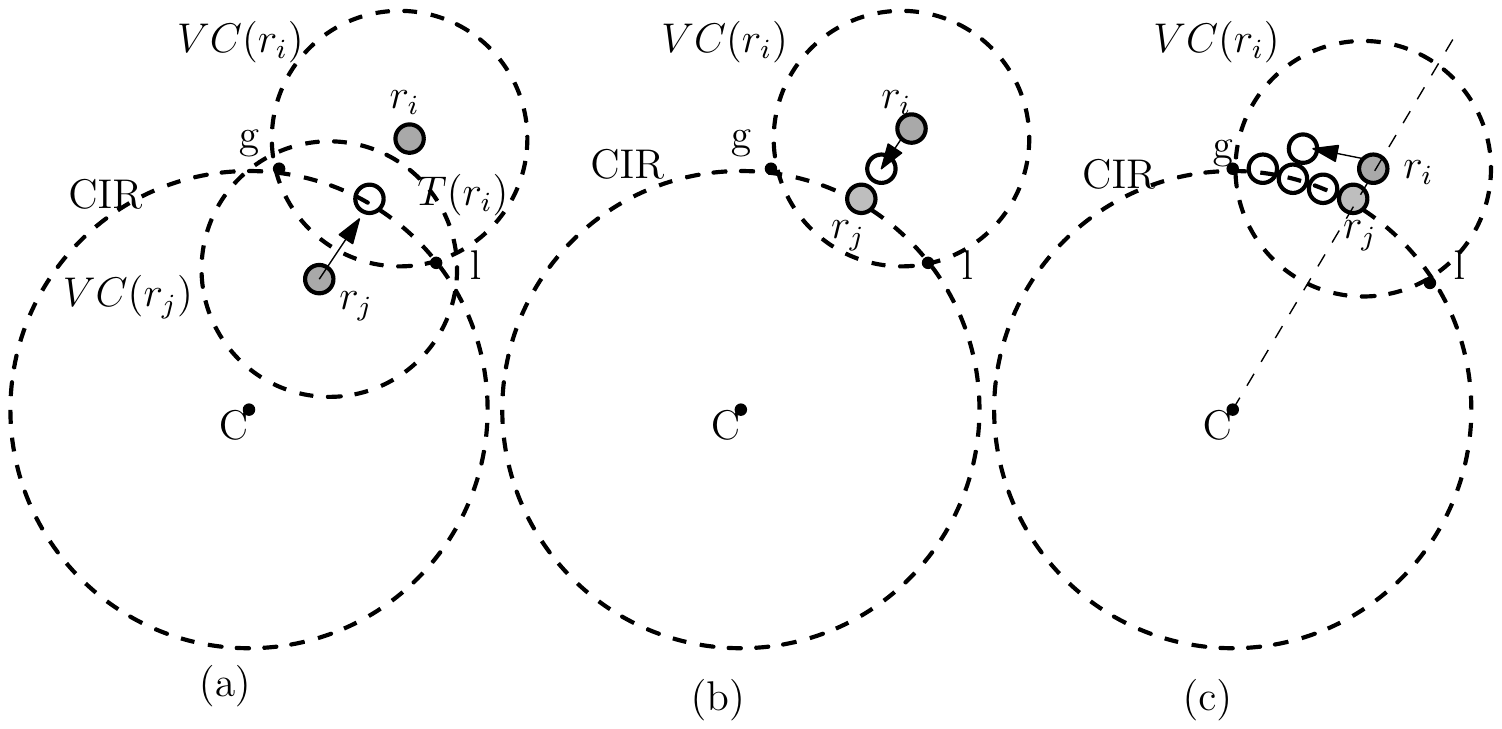} 
        \caption{An example of the configuration $\Psi_{9}$(ii)}
        \label{ii_15}
    \end{figure}
\end{itemize}

\begin{algorithm}[H]
\caption{ComputeDestination(R)}
\KwIn{(i) $r_{i} \in$ R}
\KwOut{The destination for $r_{i}$, $T(r_{i})$}
    $g \leftarrow$  point where $VC(r_{i})$ intersects CIR at left of $r_{i}$; 
        $l \leftarrow$  point where $VC(r_{i})$ intersects CIR at right of $r_{i}$;\\
        $t \leftarrow$  projpt(r,C);\\
\eIf{$r_{i}$ is in configuration $\Psi_{0}$}  
{       $r_{i}$ moves radially outward to the available vacant space;
}
{ $r_{i}$ does not move until vacant position is available outside CIR}

If $r_{i}$ is in configuration $\Psi_{1}$ (Figure \ref{ii_6}) then $r_{i}$ does not move;

\If{$r_{i}$ is in configuration $\Psi_{2}$ (Figure \ref{ii_7})}
{        h $\leftarrow$ point where $VC(r_{i})$ touches CIR;\\
        \eIf{h is vacant}
        {
           $T(r_{i}) \leftarrow$ h; (Figure \ref{ii_7}(a))
        }
        {
            $T(r_{i}) \leftarrow$ midpoint of the line joining $r_{i}$ and h; (Figure \ref{ii_7}(b))
        }
}
\If{$r_{i}$ is in configuration $\Psi_{3}$ (Figure \ref{ii_8})}
{        ti $\leftarrow$  projpt($r_{i}$, $VC(r_{i})$);\\
        \eIf {t is vacant}
        {
            $T(r_{i} \leftarrow$ ti; (Figure \ref{ii_8}(a))
        }
        {
            $T(ri) \leftarrow$  midpoint of line joining $r_{i}$ and ti; (Figure \ref{ii_8}(b))
        }
}

\end{algorithm}

\begin{algorithm}[H]               
\caption {ComputeDestination(R) continued}
\If{$r_{i}$ is in configuration $\Psi_{4}$ (Figure \ref{ii_9})}
{       m $\leftarrow$ Intersection point of the positive X-axis of robot $r_{i}$ and $VC(r_{i})$;\\
        \eIf{m is vacant}
        {
            $T(r_{i}) \leftarrow$ m; (Figure \ref{ii_9}(a))
        }
        {
            $T(r_{i}) \leftarrow$  midpoint of the line joining $r_{i}$ and m; (Figure \ref{ii_9}(b))
        }
}

\If {$r_{i}$ is in configuration $\Psi_{5}$ (Figure \ref{ii_10})}
{   \eIf {$\exists$ vacant t on the radially outward projpt(r,C) of the arc gl}
    {
        $T(r_{i}) \leftarrow$  t; (Figure \ref{ii_10}(a))
    }
    {check for the next target point on the right of it up to l.}
    
    \eIf{$\exists$ no t on the radially outward projpt(r,C) but there is t on the arc gl}
    {
        $T(r_{i}) \leftarrow$  t; (Figure \ref{ii_10}(b))
    }
    {check for the next target point on the right of it up to l.}
    
    \If{$\exists$ no vacant target position t at all on the arc tl}
    {
    \If {next position on right of the same concentric circle C(m-1) is vacant} 
    {  $r_{i}$ on C(m-1) moves rightwards to the next position\\
            If $T(r_{i})$ found on C(m) then  
            move to T(ri) and exit$\Psi_{5}$;\\
	         If $T(r_{i})$ not found then repeat $\Psi_{5}$(c);  (Figure \ref{ii_10}(c))
	  }
	  \If {next position on right of same concentric circle C(m-1) is not vacant}
      {	    $r_{i}i$ on C(m-1) does not move until the robot on next right position of $r_{i}$, has moved to C(m).
      }
    }
}

\If{$r_{i}$ is in configuration $\Psi_{6}$ (Figure \ref{ii_11})}
{        h $\leftarrow$ point where $VC(r_{i})$ touches CIR;\\
        \eIf{h is vacant}
        {
           $T(r_{i}) \leftarrow$ h; (Figure \ref{ii_11}(a))
        }
        {
            $T(r_{i}) \leftarrow$ midpoint of the line joining $r_{i}$ and h; (Figure \ref{ii_11}(b))
        }
}

\end{algorithm}

\begin{algorithm}[H]               
\caption {ComputeDestination(R) continued}

\If{$r_{i}$ is in configuration $\Psi_{7}$ (Figure \ref{ii_12})}
{        ti $\leftarrow$  projpt($r_{i}$, $VC(r_{i})$);\\
        \eIf {ti is vacant}
        {
            $T(r_{i} \leftarrow$ ti; (Figure \ref{ii_12}(a))
        }
        {
            $T(r_{i}) \leftarrow$  midpoint of line joining $r_{i}$ and ti; (Figure \ref{ii_12}(b))
        }
}

\If {$r_{i}$ is in $\Psi_{8}$ (Figure \ref{ii_13})}
{    
    \eIf{$\exists$ vacant t on radially inward projpt(r,C) of arc gl}
    {
        $T(r_{i}) \leftarrow$  t; (Figure \ref{ii_13}(a)) 
    }
    {check for the next target position on the right of it up to g.}
    \eIf{$\exists$ no t on the radially inward projpt(r,C) but there is t on the arc gl}
    {
        $T(r_{i}) \leftarrow$  t; (Figure \ref{ii_13}(b))
    }
    {check for next target point on the right of it upto g.}
    
    \If{$\exists$ no vacant target position t at all on the arc gl}
    {
	    If next position on right of C(m+1) is vacant then $r_{i}$ on C(m+1) moves rightwards to next position;\\
        If $T(r_{i})$ found on C(m) then move to T(ri) and exit;\\
	   If $T(r_{i})$ not found then repeat as in $\Psi_{8}$(c); (Figure \ref{ii_13}(c)\\
	    If next position on right of C(m+1) is not vacant then $r_{i}$ on C(m-1) does not move until the robot on next right position of $r_{i}$ has moved to C(m)
	}
}
\If {$r_{i}$ is in configuration $\Psi_{9}$ (Figure \ref{ii_14},\ref{ii_15})}
{    \If {$r_{i}$ is in $\Psi_{9}(i)$ (Figure \ref{ii_14})}
        {   $r_{i}$ moves radially inwards and $r_{j}$ moves radially outwards towards the CIR till the target point $T$ is visible to both $r_{i}$ and $r_{j}$; \\
        $r_{j}$ moves to target position $T$;
        $r_{i}$ will move to the next vacant target position on its right.
        }
    \If {$r_{i}$ is in $\Psi_{9}(ii)$ (Figure \ref{ii_15})}
        {
	    $r_{j}$ moves radially outwards towards the CIR and occupy the vacant target position, $T$, on the CIR. 
	    \If {$T(r_{i})$ is not vacant} {$r_{j}$ moves as in configuration $\Psi_{8}$(c).
	    $r_{i}$ will move to the next available vacant target position on its right side, on the CIR as in configuration $\Psi_{8}$(c)}
	    }
}

return $T(r_{i})$;

\end{algorithm}

\textbf{Correctness of ComputeDestination(R):}
The following lemmas and observations prove the correctness of the algorithm ComputeDestination(R).

\vspace{1mm}
\textbf{\textit{Observation 1:}}$r_{i}$ never goes outside of VC($r_{i}$).

\vspace{1mm}
\textbf{\textit{Observation 2:}} If two robots $r_{i}$ and $r_{j}$ are in $\Phi$1, their movements are not affected by each other.

\textbf{\textit {Proof:}} If $r_{i}$ and $r_{j}$ are in $\Phi$1, then $r_{i}$ and $r_{j}$ can not see each other. Following constraints 1 and 2, $r_{i}$ and $r_{j}$ execute the algorithm and find deterministic destinations of their own. Hence, the movements of $r_{i}$ and $r_{j}$ are not affected by each other.

\vspace{1mm}
\textbf{\textit{Observation 3:}} If two robots $r_{i}$ and $r_{j}$ are in $\Phi$2 (Figure \ref{ii_4}), and there is a robot (say $r_{k}$) at the touching point of VC($r_{i}$) and VC($r_{j}$) ($r_{k}$ is visible by both $r_{i}$ and $r_{j}$ ), then the movements of $r_{k}$, $r_{i}$ and $r_{j}$ are not affected by each other.

\textbf{Proof:} VC($r_{i}$) and VC($r_{j}$) touch at one point. Let $r_{k}$ be at the touching point. $r_{k}$ is on the straight line joining $r_{i}$ and $r_{j}$ . $r_{k}$ can see both $r_{i}$ and $r_{j}$ . $r_{i}$ and $r_{j}$ both can see $r_{k}$. However $r_{i}$ and $r_{j}$ are not mutually visible.
Let us consider the case in Figure \ref{ii_4}(a). Here dist(C; $r_{i}$) = dist(C; $r_{j}$) and dist(C; $r_{i}$); dist(C; $r_{j}$) $>$ dist(C; $r_{k}$). Following constraint 1, $r_{k}$ will not move.
However, both $r_{i}$ and $r_{j}$ will move to their deterministic destinations considering constraints 1 and 2. Let us consider the case in Figure \ref{ii_4}(b). Here $r_{j}$ , $r_{k}$ and $r_{i}$ will move one by one following constraints 1 and 2. Hence, the movements of $r_{k}$, $r_{i}$ and $r_{j}$ are not affected by each other.

\vspace{1mm}
\textbf{\textit{Observation 4:}} If two robots $r_{i}$ and $r_{j}$ are in $\Phi$3 (Figure \ref{ii_5} $\Phi$3), and there is any robot (say $r_{k}$) in common visible region ($r_{k}$ is visible by both $r_{i}$ and $r_{j}$ ), namely $\Delta$, the movements of $r_{k}$, $r_{i}$ and $r_{j}$ are not affected by each other.

\textbf{Proof:} VC($r_{i}$) and VC($r_{j}$) intersect each other and there is a robot $r_{k}$ in the common visibility region $\Delta$. $r_{k}$ can see both $r_{i}$ and $r_{j}$ . And $r_{i}$ and $r_{j}$ can see $r_{k}$ but $r_{i}$ and $r_{j}$ can not see each other (Figure \ref{ii_5} $\Phi$3).
If $r_{k}$ is on the line joining $r_{i}$ and $r_{j}$ or below the line in $\Delta$. Following constraint 1, $r_{k}$ will not move. $r_{i}$ and $r_{j}$ will move to their deterministic destinations using constraints 1 and 2.
If $r_{k}$ is above the line joining $r_{i}$ and $r_{j}$ in $\Delta$, then following constraint 1, $r_{k}$ is eligible to move and $r_{i}$ or $r_{j}$ will not move.

\vspace{1mm}
\textbf{\textit{Observation 5:}} If two robots $r_{i}$ and $r_{j}$ are in $\Phi$4 (Figure \ref{ii_5} $\Phi$4), (they are mutually visible) and there is any robot (say $r_{k}$) in common visible region $\Delta$ ($r_{k}$ is visible by both $r_{i}$ and $r_{j}$ ), then the movements of $r_{k}$, $r_{i}$ and $r_{j}$ are not affected by each other.

\textbf{Proof:} VC($r_{i}$) and VC($r_{j}$) intersect each other in such a way that $r_{i}$ and $r_{j}$ are mutually visible. $r_{k}$ is in $\Delta$. Therefore, $r_{i}$, $r_{j}$ and $r_{k}$ can see each other.
Since $r_{i}$, $r_{j}$ and $r_{k}$ are mutually visible, following constraints 1 and 2, the robots will move to their deterministic destinations.

\begin{lemma} 
The destination T($r_{i}$) computed by the robot $r_{i}$ using ComputeDestination(R) is deterministic.
\end{lemma}
\textbf{Proof:} Following observations 1, 2, 3, 4 and 5.

\subsection{Description of The Algorithm UniformCircleFormation}
Each robot executes the algorithm {\bf UniformCircleFormation(R)} and places itself on the circumference of CIR in a finite number of execution of the algorithm.
In this algorithm, if $r_{i}$ is not eligible to move according to constraint 1, then $r_{i}$ does not move. Otherwise, $T(r_{i})$ is computed by  ComputeDestination(R) and $r_{i}$ moves to T($r_{i}$).

\begin{algorithm}[H]
\KwIn{(i) $r_{i} \in$ R}
\KwIn{$r_{i}$ placed on the circumference of C after a finite number of execution of the algorithm.}
    \eIf{$r_{i}$ is not eligible to move according to constraint 1}
    {
        $r_{i}$ does not move;
    }
    {
        $T(r_{i}) \leftarrow$ Compute destination($r_{i}$);
        $r_{i}$ moves to T($r_{i}$);
    }
\caption{UniformCircleFormation(R)}
\end{algorithm}

\textbf{Correctness of UniformCircleFormation(R):}

\begin{lemma}
The path of each robot is obstacle-free.
\end{lemma}
\textbf{Proof:} A robot computes its destination and move to it. Since $r_{i}$ can be in any one of the discussed configurations, it is assured in the algorithm that $r_{i}$ moves to its computed destination without obstruction. 

\begin{theorem}
UniformCircleFormation(R) forms circle CIR by R in finite time.
\end{theorem}
\textbf{Proof:} Following lemma 12 and 13 we can ascertain that a uniform circle is formed in finite time.


\section{Uniform Circle Formation by Fat Robots Under Non-Uniform Limited Visibility Ranges}

\subsection{Underlying model}
 A set of robots R deployed on 2D plane is same as considered in previous section ({\it section 4} with modification being, a robot can see up to a fixed distance around itself but this distance can be different for the robots in R, i.e., non-uniform limited visibility. The robots are unaware of visibility ranges of other robots.

\subsection{Overview of the Problem}
A set of robots R = {$r_{1}, r_{2},..,r_{n}$} is given and our objective is to form a uniform circle (denoted by CIR) of radius rad and centered at C by moving the robots from R. The assumptions are also same as in section 4 of this paper. 

\begin{figure}[H]
      \centering
   \includegraphics[scale=0.7]{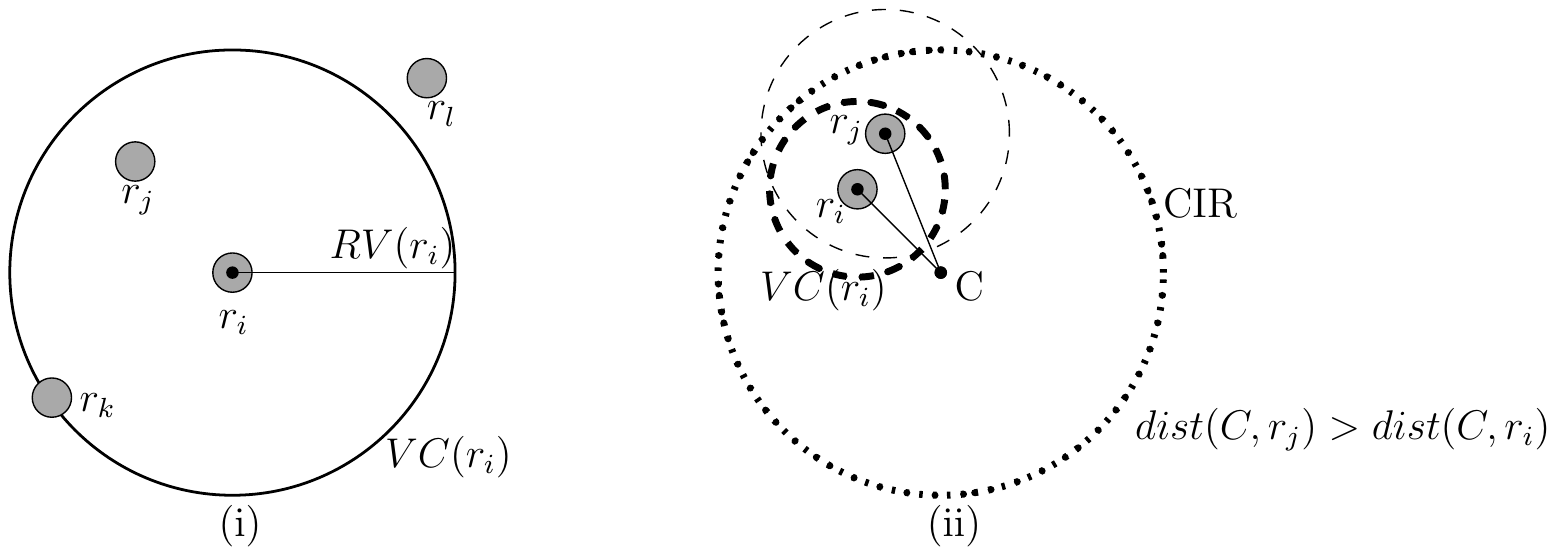} 
   \caption{(i)Visibility Circle $(VC(r_{i}))$ with Radius $(RV({r_i}))$. (ii)Example of constraint 1}
   \label{iii_1_2}
\end{figure}

\begin{definition}
Each robot can see up to a fixed distance around itself. This distance may be different for the robots and is called the visibility range of that robot. The visibility circle of $r_{i}$ in R is denoted by VC($r_{i}$ and visibility range is denoted by $R_v$ (Figure \ref{iii_1_2}(i)).
\end{definition}

\textbf{Constraints:}\\
Two same two constraints as in ({\it section 4}) have been put on the movement of any robot $r_{i}$. (Figure \ref{iii_1_2})

The algorithms ComputeTargetPoint(n,rad), ComputeRobotPosition(n,rad), ComputeDestination(R) and UniformCircleFormation(R) are executed for the formation of uniform by unit-disc robots with non-uniform visibility ranges. 

\subsection{Description of The Algorithm ComputeTargetPoint}
Let $L$ be the line parallel to the Y-axis and passing through the center $C$ of CIR) and the north-most intersection point of $L$ and CIR be $o$. $o$ is the first target point. The next target point is computed as $\frac{2\pi rad}{n}$ distance apart from $o$ at both sides of $L$. Similarly all other target points are counted such that the distance between two consecutive target points is $\frac{2\pi rad}{n}$ (Refer to section {\it section 3.4}).

\subsection{Description of The Algorithm ComputeRobotPosition}
This algorithm decides the position of robot $r_i$, either inside CIR or outside CIR or on the CIR (Refer to section {\it section 4.4}).

\subsection{Description of The Algorithm ComputeDestination}
We categorize different configurations depending on the position of visibility circles of $r_{i}$ and $r_{j}$ . We denote these configurations as $\Phi$1, $\Phi$2, $\Phi$3 and $\Phi$4.
\begin{itemize}
    \item $\Phi$1: VC($r_{i}$) and VC($r_{j}$) do not touch or intersect each other (Figure \ref{iii_3_4} $\Phi$1). 
    \item $\Phi$2: VC($r_{i}$) and VC($r_{j}$) touch each other at a single point say q.(Figure \ref{iii_3_4} $\Phi$2) If there is a robot at q say rq, then $r_{i}$ and rq and $r_{j}$ and rq are mutually visible.
        \begin{figure}[H]
            \centering
            \includegraphics[scale=0.6]{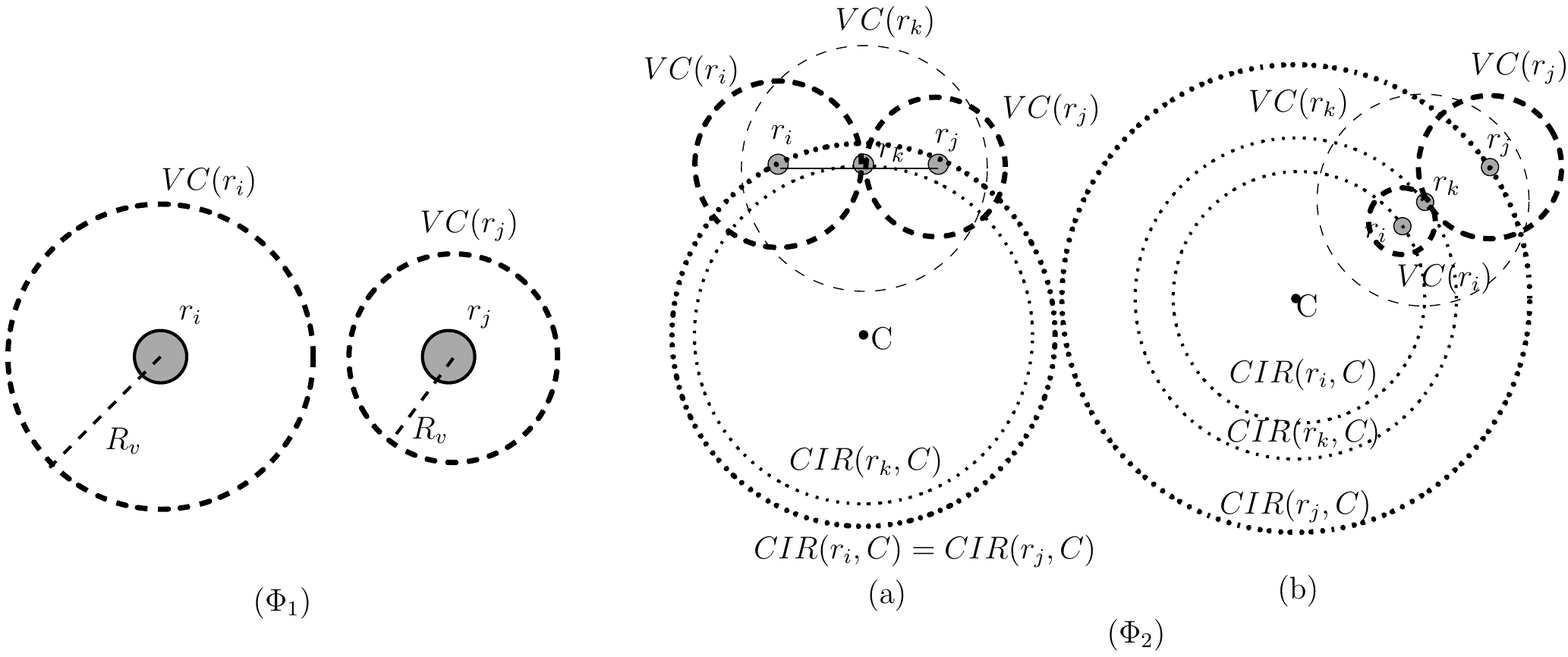} 
            \caption{Example of the configurations $\Phi$1 and $\Phi$2}
            \label{iii_3_4}
            \end{figure}

    \item $\Phi$3: VC($r_{i}$) and VC($r_{j}$) intersect each other at two points such that $r_{i}$ and $r_{j}$ can not see each other (Figure \ref{iii_5} $\Phi$3). Let $\Delta$ be the common visible region of $r_{i}$ and $r_{j}$ . If there is a robot in the region $\Delta$, say $r_{k}$, then $r_{k}$ can see $r_{i}$ and $r_{j}$ , and both $r_{i}$ and $r_{j}$ can see $r_{k}$.

    \item $\Phi$4: VC($r_{i}$) and VC($r_{j}$) intersect each other at two points such that $r_{i}$ and $r_{j}$ can see each other (Figure \ref{iii_5} $\Phi$4). Let $\Delta$ be the common visible region of $r_{i}$ and $r_{j}$ . If there is a robot in the region $\Delta$, say $r_{k}$, then $r_{k}$, $r_{i}$ and $r_{j}$ can see each other.
        \begin{figure}[H]
            \centering
            \includegraphics[scale=0.6]{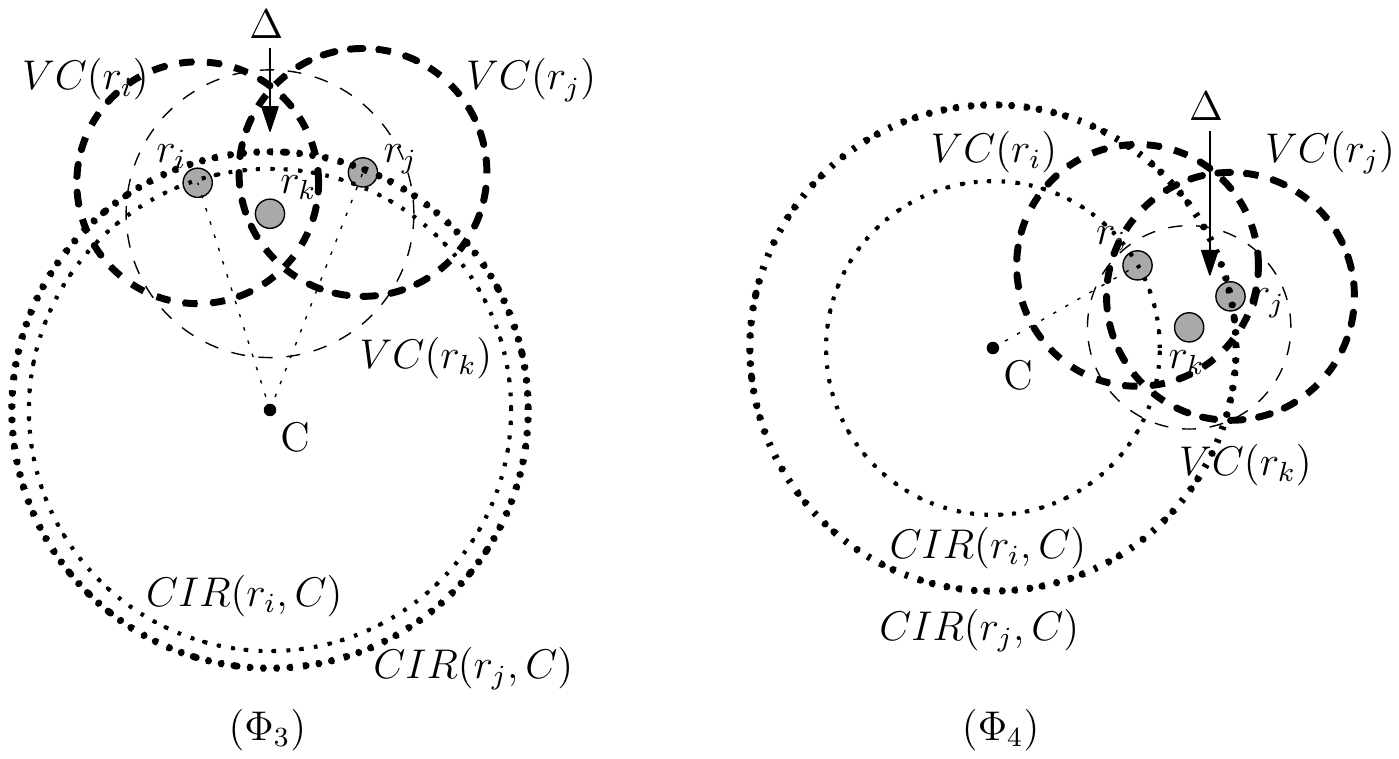} 
            \caption{Example of the configurations $\Phi$3 and $\Phi$4}
            \label{iii_5}
        \end{figure}
\end{itemize}

There are ten configurations depending on the position of $r_{i}$ inside or outside CIR. We denote these configurations as 	$\Psi_{0}$, $\Psi_{1}$, $\Psi_{2}$, $\Psi_{3}$, $\Psi_{4}$, $\Psi_{5}$, $\Psi_{6}$, $\Psi_{7}$, $\Psi_{8}$ and $\Psi_{9}$.

\begin{itemize}
    \item $\Psi_{0}$: When $r_{i}$ is on the CIR circumference and vacant space is available radially outside CIR, then $r_{i}$ moves radially outward to the available vacant space, else $r_{i}$ does not move until vacant space is available outside CIR.
    \item $\Psi_{1}$: When $r_{i}$ is on a target point, on circumference of CIR, it does not move (Figure \ref{iii_6_7} $\Psi_{1}$).
    \item $\Psi_{2}$: When $r_{i}$ is inside the CIR and VC($r_{i}$) touches CIR (at some point say h) (Figure \ref{iii_6_7} $\Psi_{2}$). If h is vacant and is a target position, then $T(r_{i})$ moves to h. Otherwise, $r_{i}$ moves to the midpoint of the line joining $r_{i}$ and h.
    \begin{figure}[H]
        \centering
        \includegraphics[scale=0.52]{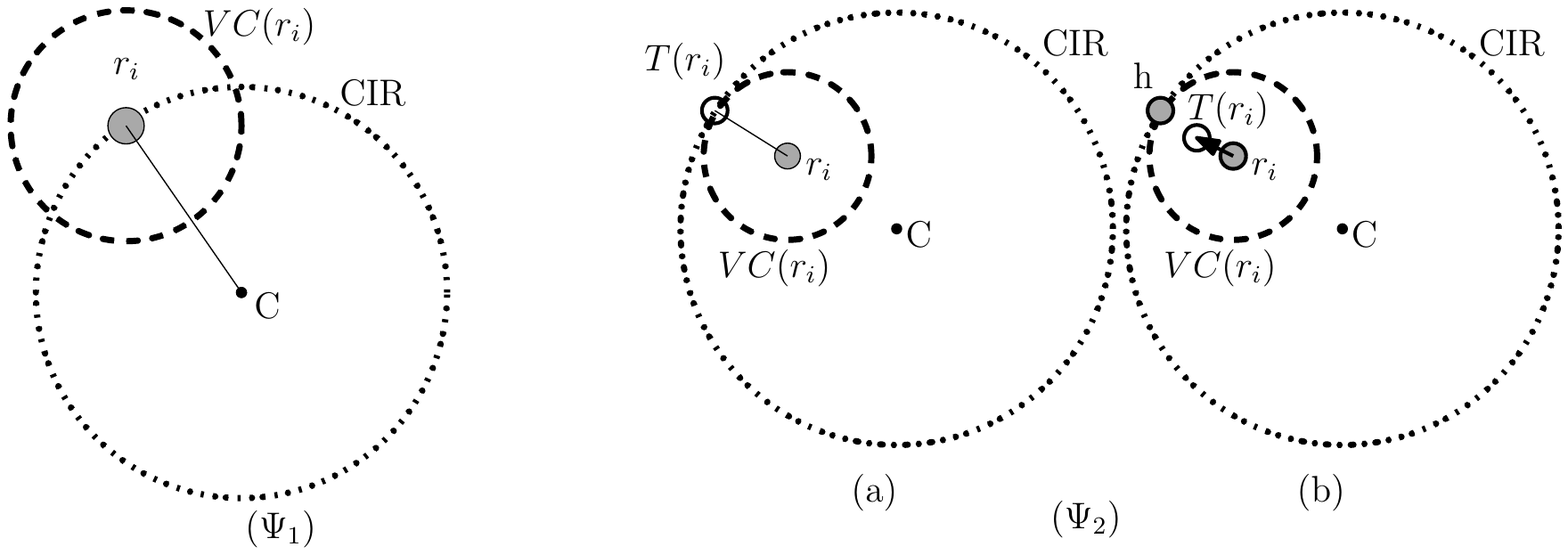} 
        \caption{Example of the configuration $\Psi_{1}$ and $\Psi_{2}$}
        \label{iii_6_7}
    \end{figure}

    \item $\Psi_{3}$: When $r_{i}$ is inside the CIR but not at C and VC($r_{i}$) does not touch or intersect the circumference of CIR (Figure \ref{iii_8_9} $\Psi_{3}$).
    Let t be  projpt($r_{i}, VC(r_{i})$, If t is a vacant point, then            $r_{i}$ moves to t. Otherwise, $r_{i}$ moves to the midpoint of the line joining $r_{i}$ and t.
    \item $\Psi_{4}$: When $r_{i}$ is at C (Figure \ref{iii_8_9} $\Psi_{4}$), it moves to the intersection point of positive X-axis of robot $r_{i} and VC(r_{i})$ (Say m). If m is a vacant point then $T(r_{i})$ moves to m, else  $T(r_{i})$ moves to the midpoint of the line joining $r_{i}$ and m.
    \begin{figure}[H]
        \centering
        \includegraphics[scale=0.61]{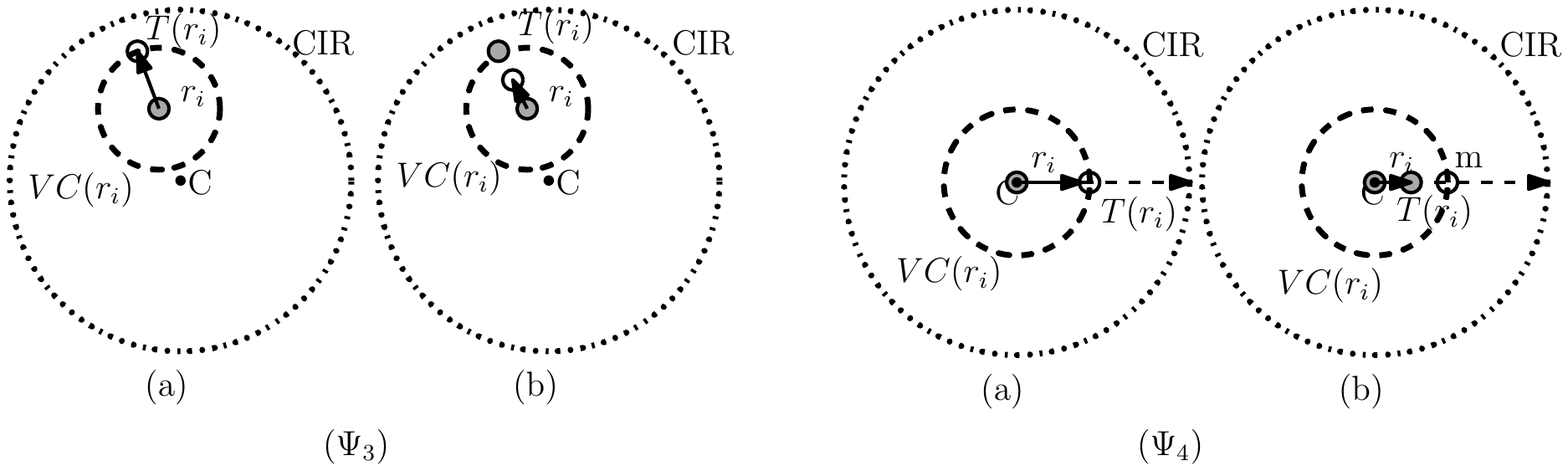} 
        \caption{Example of the configuration $\Psi_{3}$ and $\Psi_{4}$}
        \label{iii_8_9}
        \end{figure}

    \item $\Psi_{5}$: When $r_{i}$ is inside the CIR and VC($r_{i}$) intersect CIR (at two points say g and l) (Figure \ref{iii_10}). Here, we can visualize the configuration of robots as m concentric circles whose center is C.
    We consider that the robots of $\Psi_{5}$ are in concentric circle C(m-1) and they can either jump to C(m)[as in Case $\Psi_{5}$(a) and case $\Psi_{5}$(b)] or remain in the same circle C(m-1) to move rightwards [as in this case $\Psi_{5}$(c)].
	
	Possible cases for $\Psi_{5}$:
	
	\textbf{$\Psi_{5}$(a) - There is a target position $T(r_{i})$ on the radially outward projpt(r,C) of the arc gl.}
	\begin{itemize}
	    \item If $T(r_{i})$ is vacant, $r_{i}$ moves to $T(r_{i})$.
	    \item Else check for next target point on the right of it up to l.
	\end{itemize}
	
	\textbf{$\Psi_{5}$(b) - There is no target point on the radially outward projpt(r,C) but there is a target position on the arc gl.}
	\begin{itemize}
	    \item If $T(r_{i})$ is vacant, $r_{i}$ moves to $T(r_{i})$.
	    \item Else check for next target point on the right of it up to l.
	\end{itemize}
	
	\textbf{$\Psi_{5}$(c) -  There is no vacant target position at all on the arc tl.}
	\begin{itemize}
	    
	    \item If the next position on the right on the same concentric circle C(m-1) is vacant, the robot on C(m-1) moves rightwards to the next position.
	        \begin{itemize}
	            \item If T(ri) found on C(m), move to T(ri) and exit $\Psi_{5}$;
	            \item If T(ri) not found, repeat $\Psi_{5}$(c). ;
	        \end{itemize} 
	    
	    \item If the next position on the right on the same concentric circle C(m-1) is not vacant:
	    The robot, $r_{i}i$ on C(m-1) does not move until the robot on the next right position of $r_{i}$, has moved to C(m).
	   
	\end{itemize}

    \begin{figure}[H]
        \centering
        \includegraphics[scale=0.6]{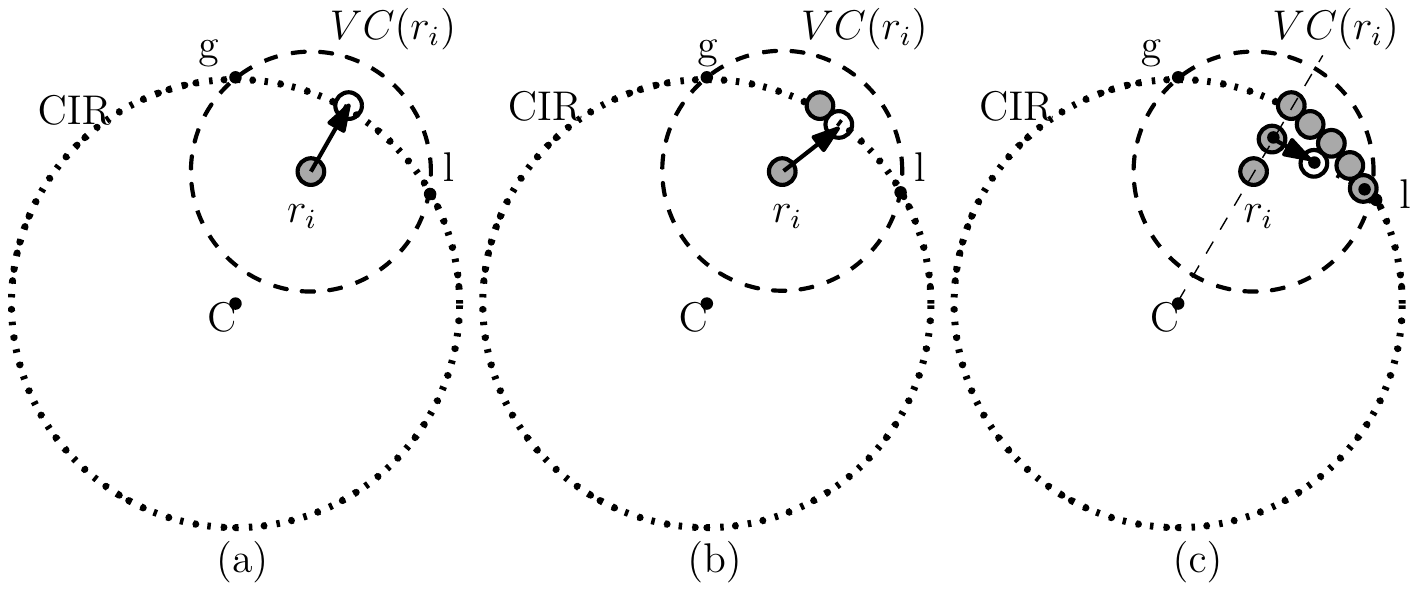} 
        \caption{Example of the configuration $\Psi_{5}$}
        \label{iii_10}
    \end{figure}

    \item $\Psi_{6}$: When $r_{i}$ is outside the CIR and VC($r_{i}$) touches CIR (at some point say h) (Figure \ref{iii_11_12} $\Psi_{6}$). If h is a vacant point, then $r_{i}$ moves to h. Otherwise, $r_{i}$ moves to the midpoint of the line joining $r_{i}$ and h.
    
    \item $\Psi_{7}$: When $r_{i}$ is outside the CIR and VC($r_{i}$) do not touch or intersect the circumference of CIR (Figure \ref{iii_11_12} $\Psi_{7}$). Let, t be projpt($r_{i}$, $VC(r_{i})$). If t is a vacant point, then $r_{i}$ moves to t. Otherwise, $r_{i}$ moves to the midpoint of the line joining $r_{i}$ and t.
    \begin{figure}[H]
        \centering
        \includegraphics[scale=0.6]{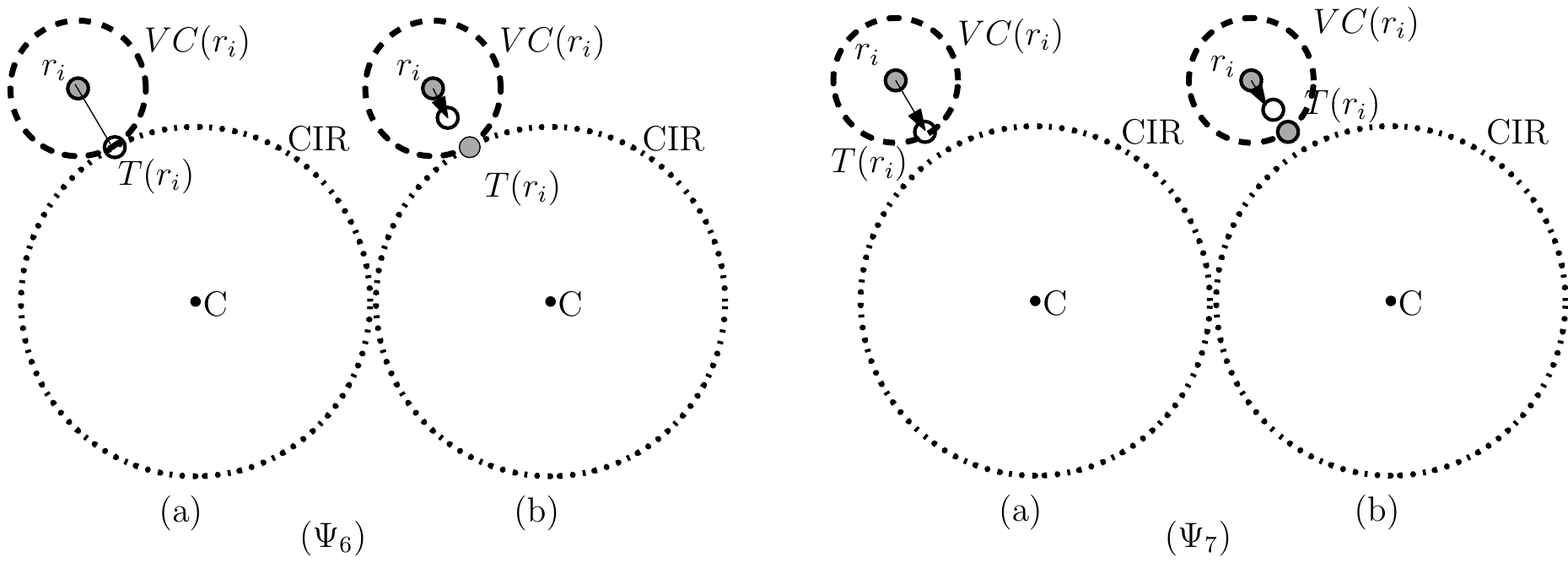}
        \caption{Example of the configuration $\Psi_{6}$ and $\Psi_{7}$}
        \label{iii_11_12}
    \end{figure}

    \item $\Psi_{8}$: When $r_{i}$ is outside the CIR and VC($r_{i}$) intersects CIR (at two points say g and l) (Figure \ref{iii_13}). Here, we can visualize the configuration of robots as m concentric circles whose center is C.
    We consider that the robots of $\Psi_{8}$ are in concentric circle C(m+1) and they can either jump down to C(m)[as in Case $\Psi_{8}$(a) and case $\Psi_{8}$(b)] or remain in the same circle C(m+1) to move rightwards [as in this case $\Psi_{8}$(c)].
	
	Possible cases for $\Psi_{8}$:
	
	\textbf{$\Psi_{8}$(a) - There is a target position $T(r_{i})$ on the radially inward projpt(r,C) of the arc gl.}
	\begin{itemize}
	    \item If $T(r_{i})$ is vacant, $r_{i}$ moves to $T(r_{i})$.
	    \item Else check for next target point on the right of it up to l.
	\end{itemize}
	
	\textbf{$\Psi_{8}$(b) - There is no target point on the radially inward projpt(r,C) but there is a target position on the arc gl.}
	\begin{itemize}
	    \item If $T(r_{i})$ is vacant, $r_{i}$ moves to $T(r_{i})$.
	    \item Else check for next target point on the right of it up to l.
	\end{itemize}

	\textbf{$\Psi_{8}$(c) -  There is no vacant target position at all on the arc tl.}
	\begin{itemize}
	    
	    \item If the next position on the right on the same concentric circle C(m+1) is vacant, the robot on C(m+1) moves rightwards to the next position.
	        \begin{itemize}
	            \item If T(ri) found on C(m), move to T(ri) and exit $\Psi_{8}$;
	            \item If T(ri) not found, repeat $\Psi_{8}$(c). ;
	        \end{itemize} 
	    
	    \item If the next position on the right on the same concentric circle C(m+1) is not vacant:
	    The robot, $r_{i}i$ on C(m+1) does not move until the robot on the next right position of $r_{i}$, has moved down to C(m).
	 \textbf{$\Psi_{8}$(c) -  There is no vacant target position at all on the arc tl.}  
	\end{itemize}

    \begin{figure}[H]
        \centering
        \includegraphics[scale=0.6]{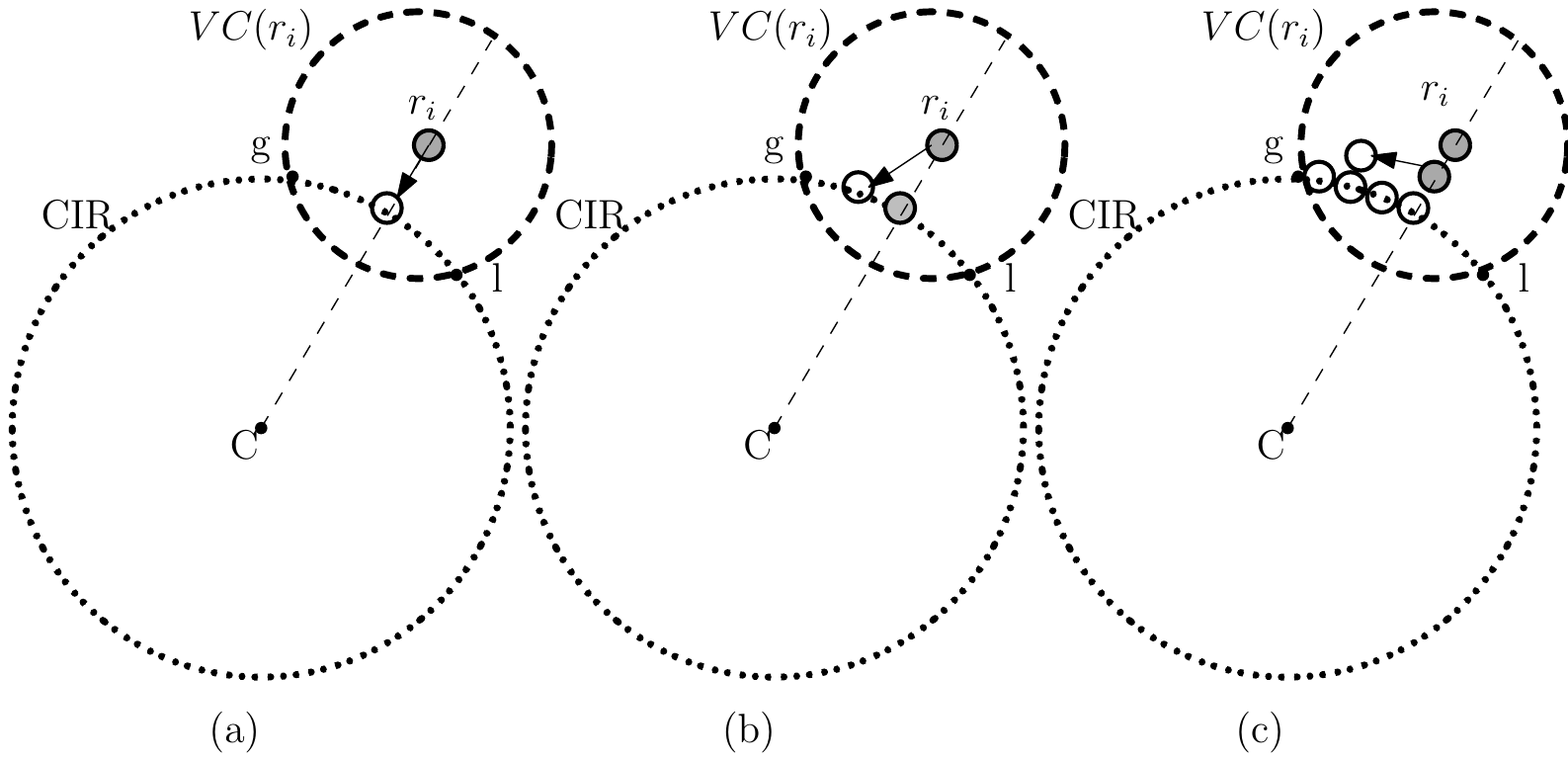} 
        \caption{An example of the configuration $\Psi_{8}$}
        \label{iii_13}
    \end{figure}
    \item $\Psi_{9}$: 
    \textbf{(i) - When $r_{i}$ is outside the CIR with VC($r_{i}$) touching CIR at a point and there is a target position $T$ at that point. Also, another robot $r_{j}$ is inside the CIR with VC($r_{j}$) touching CIR at the same point (Figure \ref{iii_14}).}
    \begin{itemize}
	    \item The robots $r_{i}$ moves radially inwards and $r_{j}$ moves radially outwards towards the CIR till the target point $T$ is visible to both $r_{i}$ and $r_{j}$.
	    \item The robot inside the CIR $r_{j}$ will move to the target position $T$ and the robot outside CIR, $r_{i}$ will move to the next vacant target position on its right.
	\end{itemize}
	\begin{figure}[H]
        \centering
        \includegraphics[scale=0.61]{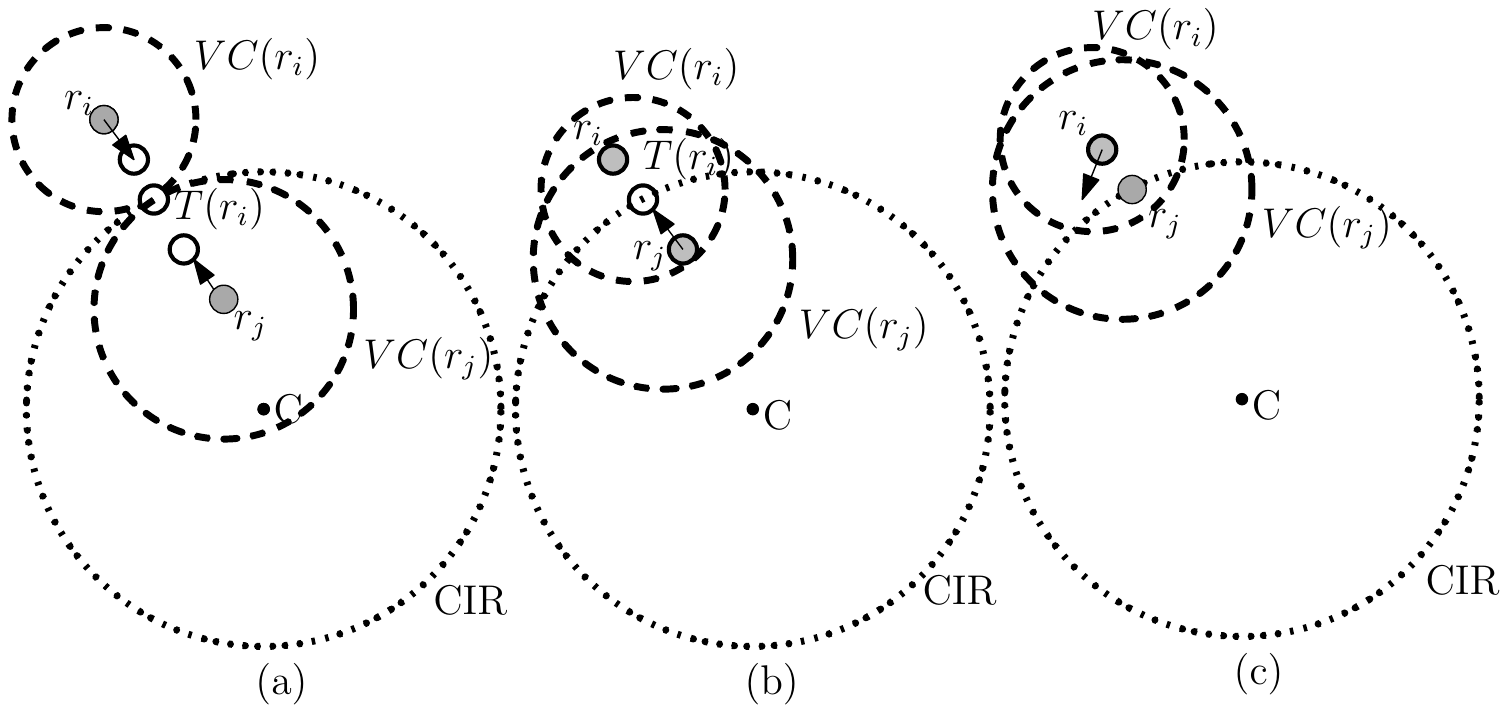} 
        \caption{An example of the configuration $\Psi_{9}$(i)}
        \label{iii_14}
    \end{figure}
    \textbf{(ii) - When $r_{i}$ is outside the CIR with VC($r_{i}$) intersecting CIR (at two points say g and l) and $r_{j}$ is inside the CIR with VC($r_{j}$) intersecting CIR (Figure \ref{iii_15}).}
    \begin{itemize}
	    \item The robot inside CIR, $r_{j}$ will move radially outwards towards the CIR and occupy the vacant target position, $T$, on the CIR.
	    If $T(r_{i})$ is not vacant $r_{j}$ moves as in configuration $\Psi_{8}$(c).
	    \item The robot outside the CIR, $r_{i}$ will move to the next available vacant target position on its right side, on the CIR as in configuration $\Psi_{8}$(c).
	\end{itemize}
    \begin{figure}[H]
        \centering
        \includegraphics[scale=0.61]{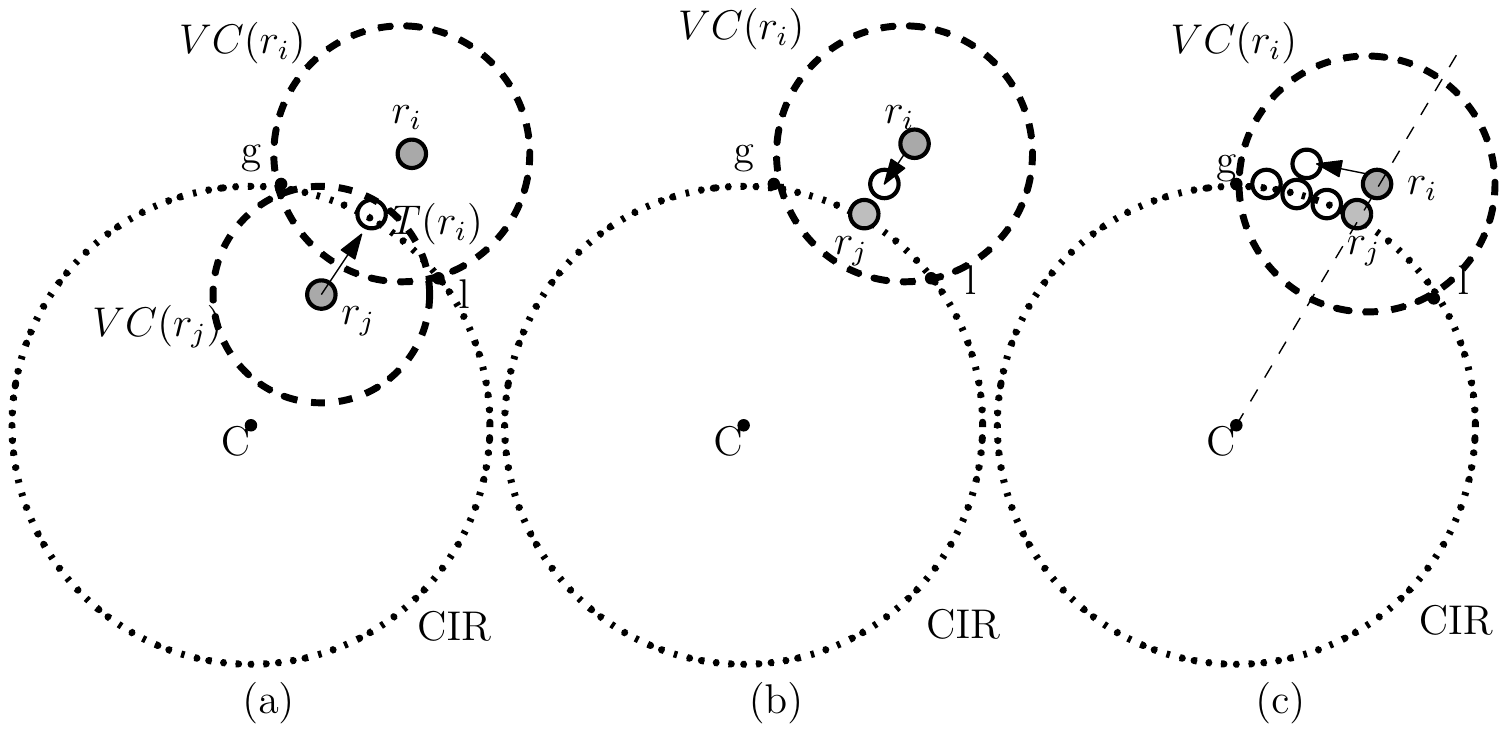} 
        \caption{An example of the configuration $\Psi_{9}$(ii)}
        \label{iii_15}
    \end{figure}
\end{itemize}

\begin{definition}{}
The destination T($r_{i}$), computed by $r_{i}$ is called a unique destination if no other robot in R computes the same destination for itself.
\end{definition}

\begin{lemma} The destination T($r_{i}$) computed by the robot $r_{i}$ using ComputeDestination(R) is unique.
\end{lemma}
\textbf{Proof:} Since a robot can be in any one of the discussed configurations, it is assured in the algorithm, while handling these configurations, that the destination T($r_{i}$) computed by the robot $r_{i}$ using ComputeDestination(R) is unique.

\subsection{Description of The Algorithm UniformCircleFormation}
Each robot executes the algorithm {\bf UniformCircleFormation(R)} and places itself on the circumference of CIR in a finite number of execution of the algorithm (Refer to section {\it section 4.6}).

\section{Conclusion}
In this paper, we have addressed three variations of the uniform circle formation problem. Collision avoidance was an important objective for the algorithms and it has been addressed successfully. 

To begin with, the first part of the paper presents an outline of a distributed algorithm that assumes on semi-synchrony, rigidity of movement and one axis agreement of the swarm robots, chosen to confirm the correctness of the algorithm. The second part of the paper presents an outline of a distributed algorithm that allows a set of autonomous, oblivious, non-communicative, asynchronous, fat robots having limited visibility to form a uniform circle executing a finite number of operation cycles. The correctness of the algorithm has been theoretically proved. In the third part of the paper, a distributed algorithm is presented for uniform circle formation by autonomous, oblivious, non-communicative, asynchronous, fat robots having non-uniform limited visibility ranges. The algorithm ensures that multiple mobile robots will form a circle of given radius and center, in a finite time and without collision. 

To the best of our knowledge, this is the first work to address these problems. Our proposed algorithms are complete and converge in finite time. In the future, we plan to consider all real-time scenarios and make the robots' model weaker. Also, we plan to add simulation results to test the feasibility of the proposed approach.

\bibliography{FormCircle}

\end{document}